\documentclass[conference]{IEEEtran}
\IEEEoverridecommandlockouts
\usepackage{cite}
\usepackage{amsmath,amssymb,amsfonts}
\usepackage{algorithm}
\usepackage{algorithmic}
\usepackage{graphicx}
\usepackage{textcomp}
\usepackage{xcolor}
\usepackage{bm}
\usepackage{booktabs}
\usepackage{multicol}
\usepackage{multirow}
\usepackage{array}
\def\BibTeX{{\rm B\kern-.05em{\sc i\kern-.025em b}\kern-.08em
    T\kern-.1667em\lower.7ex\hbox{E}\kern-.125emX}}
\begin{document}

\title{Multi-Label Learning with Label Enhancement}

\author{

\IEEEauthorblockN{Ruifeng Shao,\qquad Ning Xu,\qquad Xin Geng}
\IEEEauthorblockA{\textit{MOE Key Laboratory of Computer Network and Information Integration,} \\
\textit{School of Computer Science and Engineering, Southeast University, Nanjing 210096, China}\\
\textit{Email: \{shaorf, xning, xgeng\}@seu.edu.cn}}

}

\maketitle

\begin{abstract}
The task of multi-label learning is to predict a set of relevant labels for the unseen instance. Traditional multi-label learning algorithms treat each class label as a logical indicator of whether the corresponding label is relevant or irrelevant to the instance, i.e., +1 represents relevant to the instance and -1 represents irrelevant to the instance. Such label represented by -1 or +1 is called \emph{logical label}. Logical label cannot reflect different label importance. However, for real-world multi-label learning problems, the importance of each possible label is generally different. For the real applications, it is difficult to obtain the label importance information directly. Thus we need a method to reconstruct the essential label importance from the logical multi-label data. To solve this problem, we assume that each multi-label instance is described by a vector of latent real-valued labels, which can reflect the importance of the corresponding labels. Such label is called \emph{numerical label}. The process of reconstructing the numerical labels from the logical multi-label data via utilizing the logical label information and the topological structure in the feature space is called \emph{Label Enhancement}. In this paper, we propose a novel multi-label learning framework called LEMLL, i.e., \emph{Label Enhanced Multi-Label Learning}, which incorporates regression of the numerical labels and label enhancement into a unified framework. Extensive comparative studies validate that the performance of multi-label learning can be improved significantly with label enhancement and LEMLL can effectively reconstruct latent label importance information from logical multi-label data.
\end{abstract}


\section{Introduction}
In multi-label learning, each training instance is associated with multiple class labels and the task of multi-label learning is to predict a set of relevant labels for the unseen instance. During the past years, multi-label learning techniques have been widely applied to various fields such as document classification \cite{rubin2012statistical}, video concept detection \cite{wang2011transductive}, image classification \cite{cabral2011matrix}, audio tag annotation \cite{lo2011cost}, etc.

\begin{figure}[h]
\centering\includegraphics[width=2.4in]{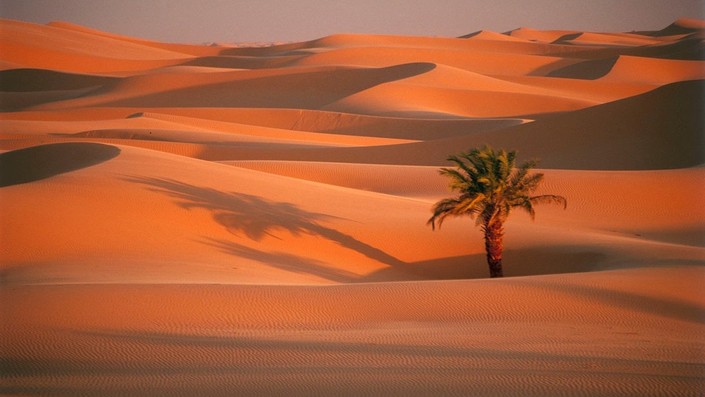}
\caption{An exemplar natural scene image which has been annotated with multiple labels \emph{sky}, \emph{desert} and \emph{tree}.}
\end{figure}

Formally speaking, let $\mathcal{X}$ = $\mathcal{R}$$^d$  be the $d$-dimensional feature space and $\mathcal{Y}=\{y_1, y_2, ... , y_l\}$ be the label set with $l$ possible labels. Given a multi-label training set $\mathcal{D}$ = \{($\bm{x}_i$,$\bm{y}_i$)$|$ $1 \leq i\leq n$\}, where $\bm{x}_i \in \mathcal{X}$ is the $d$-dimensional feature vector and $\bm{y}_i\in\{-1,+1\}^l$ is the label vector, the task of multi-label learning is to learn a multi-label predictor mapping from the space of feature vectors to the space of label vectors \cite{tsoumakas2010mining}. Traditional multi-label learning approaches treat each class label as a logical indicator of whether the corresponding label is relevant or irrelevant to the instance, i.e., $+1$ represents relevant to the instance and $-1$ represents irrelevant to the instance. Such label represented by $-1$ or $+1$ is called \emph{logical label}. Furthermore, traditional approaches take the common assumption of equal label importance, i.e., the relative importance between relevant labels is not differentiated \cite{zhang2014review}.

For real-world multi-label learning problems, the importance of each possible label is generally different. In detail, the difference of the label importance could be two-fold: 1) \emph{relevant label variance}, i.e., different labels relevant to the same instance have different relevant levels. 2) \emph{irrelevant label variance}, i.e., different labels irrelevant to the same instance have different irrelative levels. For example, as shown in Fig. 1 which is an image with five possible labels \emph{sky}, \emph{desert}, \emph{tree}, \emph{camel} and \emph{fish}, the logical label vector $[+1,+1,+1,-1,-1]^T$ is provided by the annotator. For the relevant label variance, the label importance of \emph{desert} should be greater than that of \emph{tree} and \emph{sky}, because \emph{desert} can describe the image more apparently. For the irrelevant label variance, the label importance of \emph{camel} should be greater than that of \emph{fish}, because although both are not shown in the image, it is obvious that \emph{fish} is more irrelevant to this picture than \emph{camel}.

As mentioned above, logical label uses $+1$ or $-1$ to describe each instance, which cannot reflect different label importance. So logical label can be viewed as a simplification of the instance's essential class description. However, for real-world applications, it is difficult to obtain the label importance information directly. Thus we need a method to reconstruct the latent label importance information from the logical multi-label data. To reconstruct the essential class description of each instance, we assume that there is a vector of latent real-valued labels to describe each multi-label instance, which can reflect the importance of the corresponding labels. Such label is called \emph{numerical label}. The process of reconstructing the numerical labels from the logical multi-label data via utilizing the logical label information and the topological structure in the feature space is called \emph{Label Enhancement} (LE).

In this paper, we propose an effective multi-label learning approach based on LE named \emph{ Label Enhanced Multi-Label Learning} (LEMLL). In our approach, we formulate the problem by incorporating regression of the numerical labels and label enhancement into a unified framework, where numerical labels and predictive model are jointly learned.


\section{Related Work}
Multi-label learning approaches can be roughly grouped into three types based on the thought of order of label correlations \cite{zhang2014review}. The simplest ones are the first-order approaches which decompose the problem into a series of binary classification problems, each for one label \cite{boutell2004learning,zhang2007ml}. The first-order approaches neglect the fact that the information of one label may be helpful for the learning of another label. The second-order approaches consider the correlations between pairs of class labels \cite{elisseeff2002kernel,furnkranz2008multilabel}. But the second-order approaches such as CLR \cite{furnkranz2008multilabel} and RankSVM \cite{elisseeff2002kernel} only focus on the difference between relevant label and irrelevant label. The high-order approaches consider the correlations among label subsets or all the class labels \cite{read2011classifier,tsoumakas2011random}. For all of them, these approaches take the equal label importance assumption. In contrast, our approach assumes that each instance is described by a vector of latent real-valued labels and the importance of the possible labels is different.

There have been some supervised learning tasks using label importance information (e.g. label distributions) as supervision information. In Label Distribution Learning (LDL) \cite{geng2016label}, the label distribution covers a number of labels, representing the degree to which each label describes the instance. Thus, the value of each label is numerical. The aim of LDL is to learn a model mapping from feature space to label distribution space. In Label Ranking (LR) \cite{furnkranz2003pairwise,dekel2004log,hullermeier2008label}, the label ranking of each instance describes different importance levels between labels. The goal of LR is to learn a function mapping from an instance space to rankings (total strict orders) over a predefined set of labels. However, the training of LDL or LR requires the availability of the label distributions or the label rankings in the training set. For the real applications, it is difficult to obtain such label importance information directly. On the contrary, LEMLL does not assume the availability of such explicit label importance information in training set. LEMLL can reconstruct the label importance information automatically from the logical multi-label data, while LR and LDL cannot preprocess logical label into numerical label explicitly. Therefore, LEMLL differs from these two existing works.

There have been some existing works which learn from multi-label data with auxiliary label importance information. According to \cite{brinker2007case}, Multi-Label Ranking (MLR) can be understood as learning a model that associates with a query input $\bm{x}$ both a ranking and a bipartition of the label set into relevant and irrelevant labels. A label ranking and a bipartition are given explicitly and accessible to the MLR algorithm. In \cite{cheng2010graded}, graded multi-label classification allows for graded membership of an instance belonging to a class label. An ordinal scale is assumed to characterize the membership degree and an ordinal grade is assigned for each label of the training example. In \cite{xu2013multi}, a full ordering is assumed to be known to rank relevant labels of the training example. In these cases, those auxiliary label importance information are explicitly given and accessible to the learning algorithm. Therefore, it is obvious that LEMLL is different from these existing works without assuming the availability of such explicit information.

Though there is no explicit definition of LE defined in existing literatures, some methods with similar function to LE have been proposed in the past years. In \cite{el2006study} and \cite{jiang2006fuzzy}, the membership degrees to the labels are constructed via fuzzy clustering \cite{keller1985fuzzy} and kernel method. However, these two methods have not been applied to multi-label learning. There have been some existing multi-label learning algorithms based on LE. According to \cite{li2015leveraging}, a label propagation procedure over the training instances is used to constitute the label distributions from the logical multi-label data. According to \cite{hou2016multi}, label manifold is explored to transfer the logical labels into real-valued labels. In \cite{zhang2018feature}, numerical labels are reconstructed by exploiting the structure of feature space via sparse reconstruction. These related works are all \emph{two-stage approaches}: the numerical labels are first reconstructed, and then the predictive model is trained according to the reconstructed labels. In the two-stage approaches, the results of model training cannot impact label enhancement. In contrast, the LEMLL method is a \emph{single-stage} learning algorithm where numerical labels and predictive model are jointly learned. Besides, the training of predictive model and the label enhancement are interrelated in LEMLL.

The contribution of this paper is to propose a single-stage learning strategy that jointly learns to reconstruct the numerical labels and train the predictive model. Comparing with those two-stage approaches, the LEMLL method has several advantages against those two-stage approaches: 1. LEMLL can reconstruct better latent label importance information than those two-stage approaches; 2. Learning process is single-stage, using label enhancement regularizers; 3. LEMLL has better predictive performance than those two-stage approaches.

\section{The LEMLL Approach}
\subsection{The LEMLL Framework}
Let $\mathcal{X}$ = $\mathcal{R}$$^d$ be the input space and the label space with \emph{l} logical labels can be expressed as $\{-1, +1\}$$^l$. The training set of multi-label learning can be described as $\mathcal{D}$ = \{($\bm{x}_{1}$, $\bm{y}_{1}$), ..., ($\bm{x}_{n}$,$\bm{y}_{n}$)\}. According to the above sections, we assume that the class description of each instance is a vector of numerical labels. We use $\bm{u}_{i}$ $\in$ $\mathcal{U}$ = $\mathcal{R}$$^l$ to denote the latent numerical label vector of the instance $\bm{x}_{i}$. To learn a model mapping from the input space to the numerical label space, i.e., $\emph{f}$ : $\mathcal{X}$$\to$$\mathcal{U}$, we assume that $\emph{f}$ is a linear model as:
\begin{equation}
\begin{split}
\bm{p}_i = \bm{\Theta}\varphi(\bm{x}_i)+\bm{b},\\
\end{split}
\end{equation}
where $\varphi(\bm{x}_i)$ is a nonlinear transformation of $\bm{x_i}$ to a higher dimensional feature space $\mathcal{R}$$^\mathcal{H}$, $\bm{\Theta} \in \mathcal{R}^{l \times \mathcal{H}}$ and $\bm{b} \in \mathcal{R}^{l\times 1}$ are the parameter matrices of the regression model, and $\bm{p}_i$ is the predicted numerical label vector.

Aiming at learning a model mapping from the input space to the numerical label space, a regression model can be trained by solving the following problem:
\begin{equation}
\mathop {\min }\limits_{\bm{\Theta}, \bm{b}, \bm{U}} \ \sum_{i=1}^nL_r(r_i) +  \emph{R},\\
\end{equation}
where $\emph{L}_{r}$ is a loss function, $\emph{R}$ denotes the regularizers, $r_i=\left\lVert \bm{\xi}_i \right\rVert_2=\sqrt{\bm{\xi}_i^T\bm{\xi}_i}$, $\bm{\xi}_i=\bm{\mu}_i-\bm{p}_i$ and $\bm{U}=[\bm{\mu}_1, ..., \bm{\mu}_n ]^T$.

To consider all dimensions into a unique restriction and yield a single support vector for all dimensions, the Vapnik $\varepsilon$-insensitive loss based on $\ell_2$-norm is used for $L_r$, i.e.,
\begin{equation}
L_r(r)=
\left\{
     \begin{array}{lr}
     0 &r<\varepsilon\\
     r^2-2r\varepsilon+\varepsilon^2 &r \geq \varepsilon,
     \end{array}
\right.
\end{equation}
which will create an insensitive zone determined by $\varepsilon$ around the estimate, i.e., the loss of $r$ less than $\varepsilon$ will be ignored. Because of the nonzero value of $\varepsilon$, the solution takes into account all outputs to construct each individual regressor. In this way, the cross-output relations are exploited. Furthermore, the regression model can return a sparse solution.

To control the complexity of the model, we define the following regularizer as:
\begin{equation}
\emph{R}_{1}(\bm{\Theta}) = \left\lVert \bm{\Theta} \right\rVert_F^2,
\end{equation}
where $\left\lVert \bm{\Theta} \right\rVert_F$ denotes the Frobenius norm of the matrix $\bm{\Theta}$.

\subsubsection{Label Enhancement Regularizers}
The information of the feature space and the logical label space should be used to reconstruct the numerical labels of each instance. Based on this, we give the following assumptions about label enhancement: 1) the numerical label should be close enough to the original label; 2) the numerical label space and the feature space should share similar local topological structure.

As mentioned above, logical label can be viewed as a simplification of numerical label. Intuitively, the original label contains some information of numerical label, so the original label cannot differ too much from the numerical label. Thus we can get the first assumption and define the following regularizer as:
\begin{equation}
\emph{R}_{2}(\bm{U}, \bm{Y}) = \left\lVert \bm{U} - \bm{Y} \right\rVert_F^2,
\end{equation}
where $\bm{Y}=[\bm{y}_1, ..., \bm{y}_n ]^T$ is the logical label matrix.

According to the smoothness assumption \cite{zhu2005semi}, \emph{the points close to each other are more likely to share a label}. We can easily infer that the points close to each other in the feature space are more likely to have similar numerical label vector. This intuition leads to the second assumption. The topological structure of the feature space can be expressed by a fully connected graph $\mathcal{G}=(\bm{V}, \bm{E}, \bm{W})$, where $\bm{V}$ is the vertex set of the training instances, i.e., $\bm{V} = \{\bm{x}_i | 1 \leq \emph{i} \leq \emph{n}\}$, $\bm{E}$ is the edge set in which $\bm{e}_{ij}$ represents the relationship between $\bm{x}_i$ and $\bm{x}_j$, and $\bm{W}$ is the weight matrix in which each element $W_{ij}$ represents the weight of the edge $\bm{e}_{ij}$. To estimate the local topological structure of the feature space, the local neighborhood information of each instance should be used to construct the graph $\mathcal{G}$. According to Local Linear Embedding (LLE) \cite{roweis2000nonlinear}, each point can be reconstructed by a linear combination of its neighbors. The approximation of the topological structure of the feature space can be obtained by solving the following problem:
\begin{equation}
\begin{split}
\mathop{\min }\limits_{\bm{W}} &\sum_{i=1}^n\Big\| \bm{x}_i - \sum_{j\not=i}W_{ij}\bm{x}_j \Big\|^2  \\s.t. &\sum_{i=1}^nW_{ij} = 1,
\end{split}
\end{equation}
where $W_{ij} = 0$ if $\bm{x}_j$ is not one of $\bm{x}_i$'s \emph{K}-nearest neighbors. $\sum_{i=1}^nW_{ij} = 1$ is constrained because of the translation invariance. Eq. (6) can be transformed into the \emph{n} quadratic programming problems:
\begin{equation}
\begin{split}
\mathop{\min}\limits_{\bm{W}_i}\quad&\bm{W}_i^T\bm{G}_i\bm{W}_i\\
s.t.\quad&\bm{1}^T\bm{W}_i=1,
\end{split}
\end{equation}
where $(\bm{G}_i)_{jk}=(\bm{x}_i-\bm{x}_j)^T(\bm{x}_i-\bm{x}_k)$. Because the feature space and the numerical label space should share similar local topological structure, we define the following regularizer as:
\begin{equation}
\emph{R}_{3}(\bm{W}, \bm{U}) = \left\lVert \bm{U} - \bm{WU} \right\rVert_F^2 = tr(\bm{U}^T\bm{M}\bm{U}),
\end{equation}
where $\bm{M}=(\bm{I} - \bm{W})^T(\bm{I} - \bm{W})$ and $\bm{I}$ is an identity matrix. For a matrix $\bm{A}$, $tr(\bm{A})$ is its trace.

By replacing \emph{R} in Eq. (2) with Eqs. (4), (5) and (8), the framework can be rewritten as:
\begin{equation}
\begin{split}
\mathop {\min }\limits_{\bm{\Theta}, \bm{b}, \bm{U}}&\sum_{i=1}^nL_r(r_i) + \alpha\left\lVert \bm{\Theta} \right\rVert_F^2 + \beta \left\lVert \bm{U} - \bm{Y} \right\rVert_F^2 + \gamma \ tr(\bm{U}^T\bm{M}\bm{U})\\
s.t.\quad&r_i=\left\lVert \bm{\xi}_i \right\rVert_2=\sqrt{\bm{\xi}_i^T\bm{\xi}_i}\\
&\bm{\xi}_i=\bm{\mu}_i-\bm{\Theta}\varphi(\bm{x}_i)-\bm{b}\\
&L_r(r)=
\left\{
     \begin{array}{lr}
     0 &r<\varepsilon\\
     r^2-2r\varepsilon+\varepsilon^2 &r \geq \varepsilon,
     \end{array}
\right.
\end{split}
\end{equation}
where $\alpha$, $\beta$ and $\gamma$ are tradeoff parameters.

\subsection{The Alternating Solution for the Optimization}
When we fix $\bm{U}$ to solve $\bm{\Theta}$ and $\bm{b}$, Eq. (9) can be rewritten as:
\begin{equation}
\begin{split}
\mathop {\min }\limits_{\bm{\Theta}, \bm{b}}\quad&\sum_{i=1}^nL_r(r_i) + \alpha\left\lVert \bm{\Theta} \right\rVert_F^2\\
s.t.\quad&r_i=\left\lVert \bm{\xi}_i \right\rVert_2=\sqrt{\bm{\xi}_i^T\bm{\xi}_i}\\
&\bm{\xi}_i=\bm{\mu}_i-\bm{\Theta}\varphi(\bm{x}_i)-\bm{b}\\
&L_r(r)=
\left\{
     \begin{array}{lr}
     0&r<\varepsilon\\
     r^2-2r\varepsilon+\varepsilon^2 &r \geq \varepsilon.
     \end{array}
\right.
\end{split}
\end{equation}
Notice that Eq. (10) is a MSVR with the Vapnik $\varepsilon$-insensitive loss based on $\ell_2$-norm \cite{tuia2011multioutput}. So $\bm{\Theta}$ and $\bm{b}$ can be optimized by training a MSVR model.

When we fix $\bm{\Theta}$ and $\bm{b}$ to solve $\bm{U}$, the objective function becomes:
\begin{equation}
\begin{split}
L(\bm{U})=\sum_{i=1}^nL_r(r_i) + \beta \left\lVert \bm{U} - \bm{Y} \right\rVert_F^2 + \gamma \ tr(\bm{U}^T\bm{M}\bm{U}).\\
\end{split}
\end{equation}

We use an iterative quasi-Newton method called Iterative Re-Weighted Least Square (IRWLS) \cite{perez2001fast,tuia2011multioutput} to minimize $L(\bm{U})$. Firstly, $L_r(r_i)$ is approximated by its first order Taylor expansion at the solution of the current $k$-th iteration, denoted by $\bm{U}^{(k)}$:
\begin{equation}
\begin{split}
L_r'(r_i)=L_r(r_i^{(k)})+\frac{dL_r(r)}{dr} \bigg|_{r_i^{(k)}} \frac{(\bm{\xi}_i^{(k)})^T}{r_i^{(k)}}(\bm{\xi}_i-\bm{\xi}_i^{(k)}),
\end{split}
\end{equation}
where $\bm{\xi}_i^{(k)}$ and $r_i^{(k)}$ are calculated from $\bm{U}^{(k)}$. Then a quadratic approximation is further constructed
\begin{equation}
\begin{split}
L_r''(r_i)&=L_r(r_i^{(k)})+\frac{dL_r(r)}{dr} \bigg|_{r_i^{(k)}} \frac{r_i^2-(r_i^{(k)})^2}{2r_i^{(k)}}\\
&=a_ir_i^2+\tau,
\end{split}
\end{equation}
where
\begin{equation}
\begin{split}
a_i=\frac{1}{2r_i^{(k)}}\frac{dL_r(r)}{dr} \bigg|_{r_i^{(k)}}=
\left\{
     \begin{array}{lr}
     0 &r_i^{(k)}<\varepsilon\\
     \frac{(r_i^{(k)}-\varepsilon)}{r_i^{(k)}} &r_i^{(k)} \geq \varepsilon,
     \end{array}
\right.
\end{split}
\end{equation}
and $\tau$ is a constant term that does not depend on $\bm{U}^{(k)}$. By substituting Eqs. (13) and (14) into Eq. (11), the  objective function becomes:
\begin{equation}
\begin{split}
&L''(\bm{U})=\sum_{i=1}^na_ir_i^2 + \beta \left\lVert \bm{U} - \bm{Y} \right\rVert_F^2 + \gamma\ tr(\bm{U}^T\bm{M}\bm{U}) + \nu\\
&=tr(\bm{\Xi}^T \bm{D_a} \bm{\Xi}) + \beta \left\lVert \bm{U} - \bm{Y} \right\rVert_F^2 + \gamma \ tr(\bm{U}^T\bm{M}\bm{U}) + \nu,\\
\end{split}
\end{equation}
where $\bm{\Xi}=[\bm{\xi}_1, ... , \bm{\xi}_n]^T=\bm{U}-\bm{P}$, $\bm{P}=[\bm{p}_1, ... , \bm{p}_n]^T$, $(\bm{D}_a)_{ij}=a_i\Delta_{ij}$ ($\Delta_{ij}$ is the Kronecker's delta function) and $\nu$ is a constant term. Furthermore, Eq. (15) can be rewritten as:
\begin{equation}
\begin{split}
&L''(\bm{U})=tr\big(\bm{U}^T(\bm{D_a}+\beta\bm{I}+\gamma \bm{M})\bm{U}\big)\\
&\quad\quad\quad\quad\quad\quad\quad -2tr\big((\bm{D_a}\bm{P}+\beta \bm{Y})\bm{U}^T\big) + \nu',
\end{split}
\end{equation}
where $\nu'$ is a constant term. The minimization of Eq. (16) can be solved by setting the derivative of the above target function with respect to $\bm{U}$ to be zero:
\begin{equation}
\frac{\partial L''(\bm{U})}{\partial \bm{U}}=2(\bm{D_a}+\beta\bm{I}+\gamma \bm{M})\bm{U}-2(\bm{D_a}\bm{P}+\beta \bm{Y})=0.
\end{equation}
Solving Eq. (17), we can get
\begin{equation}
\begin{split}
\bm{U}=(\bm{D}_a+\beta\bm{I}+\gamma\bm{M})^{-1}(\bm{D}_a\bm{P}+\beta\bm{Y}).
\end{split}
\end{equation}
The direction of Eq. (18) is used as the descending direction for the minimization of Eq. (11). The solution for the next iteration $\bm{U}^{(k+1)}$ is obtained via a line search algorithm along this direction.

\renewcommand{\algorithmicrequire}{ \textbf{Input:}} 
\renewcommand{\algorithmicensure}{ \textbf{Output:}} 
\begin{algorithm}[h]
  \caption{The LEMLL Algorithm}
  \begin{algorithmic}[1]
    \REQUIRE
      the training feature matrix $\bm{X}=[\bm{x}_1,...,\bm{x}_n]^T$ and the training label matrix $\bm{Y}$
    \ENSURE
      the numerical label matrix $\bm{U}$ and the parameter matrices $\bm{\Theta}$ and $\bm{b}$
    \STATE $\bm{U}^{(0)}\leftarrow\bm{0}$; $\emph{t}\leftarrow 1$;
    \STATE Construct $\bm{W}$ according to Eq. (7);
    \REPEAT
      \STATE Optimize $\bm{\Theta}^{(t)}$ and $\bm{b}^{(t)}$ with $\bm{U}^{(t-1)}$ according to Eq. (10);
      \STATE Update $\bm{P}^{(t)}$ according to Eq. (1);
      \STATE Update $\bm{U}^{(t)}$ via the IRWLS procedure;
      \STATE $t\leftarrow t+1$;
    \UNTIL{convergence reached}
    \STATE Return $\bm{U}$, $\bm{\Theta}$ and $\bm{b}$.
  \end{algorithmic}
\end{algorithm}

The pseudo code of the LEMLL algorithm is presented in \textbf{Algorithm 1}. In order to distinguish the relevant and irrelevant labels, numerical labels should be divided into two sets, i.e., the relevant and irrelevant sets. According to \cite{furnkranz2008multilabel} and \cite{li2015leveraging}, an extra virtual label $y_0$ is added into the original label set, i.e., the extended original label set $\mathcal{Y}' = \mathcal{Y} \cup \{y_0\} = \{y_0,y_1,...,y_l\}$. In this paper, the logical value of $y_0$ is set to 0. Using the extended original label set to do the training process, the optimal parameter matrices $\bm{\Theta}^{*} \in \mathcal{R}^{(l+1) \times \mathcal{H}}$ and $\bm{b}^{*} \in \mathcal{R}^{(l+1)\times 1}$ are learnt. Given a test instance $\bm{x}$, the model can predict an extended numerical label vector $\bm{p}^*$. The predicted numerical label greater than $p_0^{*}$ is relevant to the example and the label smaller than $p_0^{*}$ is irrelevant to the example.

\begin{table*}[!htb]
\caption{Characteristics of the 15 benchmark multi-label data sets used in the experiments.}
\fontsize{7}{7}\selectfont
\center
\begin{tabular}{c c c c c c c c c c}
\midrule\midrule
Data set & $|S|$ & $dim(S)$ & $L(S)$ & $F(S)$ & $LCard(S)$ & $LDen(S)$ & $DL(S)$ & $PDL(S)$ & Domain \\
\midrule
cal500 &502 &68 &174 &numeric &26.044 &0.150 &502 &1.000 &audio\\
medical &978 &1,449 &45 &nominal &1.245 &0.028 &94 &0.096 &text\\
llog &1,460 &1,004 &75 &nominal &1.180 &0.016 &304 &0.208 &text\\
enron &1,702 &1,001 &53 &nominal &3.378 &0.064 &753 &0.442 &text\\
msra &1,868 &898 &19 &numeric &6.315 &0.332 &947 &0.507 &images \\
scene &2,407 &294 &5 &numeric &1.074 &0.179 &15 &0.006 &images\\
yeast &2,417 &103 &14 &numeric &4.237 &0.303 &198 &0.082 &biology \\
slashdot &3,782 &1,079 &22 &nominal &1.181 &0.054 &156 &0.041 &text \\
corel5k &5,000 &499 &374 &nominal &3.522 &0.009 &3,175 &0.635 &images \\
rcv1-s1 &6,000 &944 &101 &numeric &2.880 &0.029 &1,028 &0.171 &text \\
rcv1-s2 &6,000 &944 &101 &numeric &2.634 &0.026 &954 &0.159 &text \\
bibtex &7,395 &1,836 &159 &nominal &2.402 &0.015 &2,856 &0.386 &text\\
corel16k-s1 &13,766 &500 &153 &nominal &2.859 &0.019 &4,803 &0.349 &images \\
corel16k-s2 &13,761 &500 &164 &nominal &2.882 &0.018 &4,868 &0.354 &images \\
tmc2007 &28,696 &981 &22 &nominal &2.158 &0.098 &1341 &0.047 &text\\
\midrule\midrule
\end{tabular}
\end{table*}

\section{Experiments}
This section is divided into two parts. In the first part, we evaluate the predictive performance of our method on multi-label data sets. In the second part, we reconstruct the label importance information from the logical labels via the LE methods, and then compare the recovered label importance with the ground-truth label importance.

\subsection{Predictive Performance Evaluation}
\subsubsection{Experimental Settings}
For comprehensive performance evaluation, a total of fifteen benchmark multi-label data sets in Mulan \cite{tsoumakas2011mulan} and Meka \cite{MEKA} are collected for experimental studies. For a data set $S$, we use $|S|$, $dim(S)$, $L(S)$, $F(S)$, $LCard(S)$, $LDen(S)$, $DL(S)$ and $PDL(S)$ to represent its number of examples, number of features, number of class labels, feature type, label cardinality, label density, distinct label set and proportion of distinct label sets respectively. Table I summarizes the characteristics of the fifteen data sets.

To examine the effectiveness of label enhancement, LEMLL is first compared with MSVR \cite{tuia2011multioutput}, which can be considered as a degenerated version of LEMLL without label enhancement. Besides, three well-established two-stage approaches are employed for comparative studies, each implemented with parameter setup suggested in respective literatures: 1) Multi-label Learning with Feature-induced labeling information Enrichment (MLFE) \cite{zhang2018feature}: [suggested setup: $\rho=1$, $c_1=1$, $c_1=2$, $\beta$$_1$, $\beta$$_2$ and $\beta$$_3$ chosen among \{1,2, ... ,10\}, \{1,10,15\} and \{1,10\} respectively ]; 2) Multi-Label Manifold Learning (ML$^2$) \cite{hou2016multi}: [suggested setup: $K$ = $l$ +1, $\lambda$ = 1, $C_{1}$ and $C_{2}$ chosen among \{ 1, 2, ... , 10 \}]; 3) RElative Labeling-Importance Aware multi-laBel learning (RELIAB) \cite{li2015leveraging}: [suggested setup: $\alpha$ = 0.5, $\beta$ chosen among \{ 0.001, 0.01, ... , 10 \}, $\tau$ chosen among \{0.1, 0.15, ... ,0.5\} ]. Besides, we choose to compare the performance of LEMLL against three state-of-the-art algorithms, including one first-order approach ML-kNN \cite{zhang2007ml}, one second-order approach Calibrated Label Ranking (CLR) \cite{furnkranz2008multilabel}, and one high-order approach Ensemble of Classifier Chains (ECC) \cite{read2011classifier}. For the three comparing algorithms, parameter configurations suggested in the literatures are used. For ML-kNN, $k$ is set to 10. The ensemble size of ECC is set to 30. The three state-of-the-art comparing algorithms are implemented under the Mulan multi-label learning package \cite{tsoumakas2011mulan} by instantiating the base learners of CLR and ECC with logistic regression. For LEMLL, \emph{K} is set to 10. $\varepsilon$ is set to 0.1. $\alpha$, $\beta$ and $\gamma$ are all chosen among \{$\frac{1}{64}$, $\frac{1}{16}$, $\frac{1}{4}$, $1$, $4$, $16$, $64$\} with cross-validation on the training set. For the sake of fairness, linear kernel is used in MSVR, ML$^2$ and LEMLL.

Five widely-used evaluation metrics are used in comparative studies: \emph{Hamming loss} (HL), \emph{Ranking loss} (RL), \emph{One-error} (OE), \emph{Coverage} (CO) and \emph{Average precision} (AP). Note that for all the five multi-label metrics, their values vary between $[ 0, 1 ]$. Furthermore, for average precision, the larger the values the better the performance; While for the other four metrics, the smaller the values the better the performance. These metrics serve as good indicators for comparative studies as they evaluate the performance of the models from various aspects. Concrete metric definitions can be found in \cite{zhang2014review}.

\renewcommand{\multirowsetup}{\centering}
\begin{table*}[!htb]
\centering
  \fontsize{7}{7}\selectfont
  \caption{Predictive performance (mean $\pm$ std. deviation). $\bullet$($\circ$) indicates LEMLL is significantly better (worse) than the corresponding method on the criterion based on paired \emph{t}-test at 95$\%$ significance level. $\downarrow$ ($\uparrow$) implies the smaller (larger), the better.}
    \begin{tabular}{ccccccccc}
    \midrule
    \midrule
    \multirow{2}{1.5cm}{Data set}&
    \multicolumn{8}{c}{Hamming loss$\downarrow$}\cr
    \cmidrule(lr){2-9}
    &LEMLL&MLFE&ML$^2$&RELIAB&MSVR&ML-kNN&CLR&ECC\cr
    \midrule
    CAL500	&0.137$\pm$0.002	&0.141$\pm$0.002$\bullet$	&0.162$\pm$0.013$\bullet$	&0.167$\pm$0.004$\bullet$	&0.137$\pm$0.002	        &0.139$\pm$0.001$\bullet$
    &0.165$\pm$0.005$\bullet$	&0.147$\pm$0.002$\bullet$	\cr
    medical	&0.011$\pm$0.001	&0.014$\pm$0.001$\bullet$	&0.019$\pm$0.003$\bullet$	&0.017$\pm$0.001$\bullet$	&0.012$\pm$0.001$\bullet$	&0.017$\pm$0.001$\bullet$
   &0.023$\pm$0.002$\bullet$	&0.013$\pm$0.001$\bullet$	\cr
    llog	&0.015$\pm$0.000	&0.021$\pm$0.001$\bullet$	&0.028$\pm$0.001$\bullet$	&0.016$\pm$0.000$\bullet$	&0.029$\pm$0.001$\bullet$	&0.015$\pm$0.000
     &0.021$\pm$0.003$\bullet$	&0.016$\pm$0.000$\bullet$	\cr
    enron	&0.050$\pm$0.001	&0.116$\pm$0.005$\bullet$	&0.160$\pm$0.015$\bullet$	&0.062$\pm$0.003$\bullet$	&0.070$\pm$0.001$\bullet$	&0.055$\pm$0.001$\bullet$
    &0.072$\pm$0.002$\bullet$	&0.064$\pm$0.001$\bullet$	\cr
    msra	&0.187$\pm$0.009	&0.211$\pm$0.006$\bullet$	&0.221$\pm$0.005$\bullet$	&0.279$\pm$0.018$\bullet$	&0.193$\pm$0.008	        &0.213$\pm$0.008$\bullet$
      &0.342$\pm$0.035$\bullet$	&0.353$\pm$0.039$\bullet$	\cr
    scene	&0.109$\pm$0.003	&0.125$\pm$0.002$\bullet$	&0.140$\pm$0.012$\bullet$	&0.127$\pm$0.005$\bullet$	&0.115$\pm$0.003$\bullet$	&0.092$\pm$0.003$\circ$
        &0.181$\pm$0.004$\bullet$	&0.133$\pm$0.002$\bullet$	\cr
    yeast	&0.201$\pm$0.002	&0.227$\pm$0.003$\bullet$	&0.230$\pm$0.003$\bullet$	&0.214$\pm$0.004$\bullet$	&0.204$\pm$0.003$\bullet$	&0.206$\pm$0.001$\bullet$
        &0.222$\pm$0.003$\bullet$	&0.216$\pm$0.002$\bullet$	\cr
    slashdot	&0.041$\pm$0.001	&0.074$\pm$0.002$\bullet$	&0.050$\pm$0.001$\bullet$	&0.060$\pm$0.002$\bullet$	&0.043$\pm$0.001$\bullet$	&0.052$\pm$0.000$\bullet$
     &0.058$\pm$0.001$\bullet$	&0.049$\pm$0.001$\bullet$	\cr

    \midrule
    \multirow{2}{1.5cm}{Data set}&
    \multicolumn{8}{c}{Ranking loss$\downarrow$}\cr
    \cmidrule(lr){2-9}
    &LEMLL&MLFE&ML$^2$&RELIAB&MSVR&ML-kNN&CLR&ECC\cr
    \midrule
    CAL500	&0.182$\pm$0.003	&0.185$\pm$0.003	        &0.209$\pm$0.019$\bullet$	&0.181$\pm$0.003	        &0.182$\pm$0.003	        &0.189$\pm$0.002$\bullet$
  &0.239$\pm$0.028$\bullet$	&0.205$\pm$0.004$\bullet$	\cr
    medical	&0.027$\pm$0.005	&0.035$\pm$0.004$\bullet$	&0.059$\pm$0.008$\bullet$	&0.034$\pm$0.006$\bullet$	&0.040$\pm$0.008$\bullet$	&0.055$\pm$0.007$\bullet$
   &0.123$\pm$0.028$\bullet$	&0.032$\pm$0.007	\cr
    llog	&0.146$\pm$0.008	&0.266$\pm$0.013$\bullet$	&0.310$\pm$0.010$\bullet$	&0.124$\pm$0.005$\circ$	    &0.307$\pm$0.009$\bullet$	&0.168$\pm$0.007$\bullet$
   &0.197$\pm$0.018$\bullet$	&0.154$\pm$0.009	\cr
    enron	&0.084$\pm$0.003	&0.200$\pm$0.007$\bullet$	&0.271$\pm$0.019$\bullet$	&0.091$\pm$0.003$\bullet$	&0.189$\pm$0.006$\bullet$	&0.100$\pm$0.002$\bullet$
    &0.089$\pm$0.002$\bullet$	&0.120$\pm$0.004$\bullet$	\cr
    msra	&0.134$\pm$0.011	&0.161$\pm$0.007$\bullet$	&0.167$\pm$0.007$\bullet$	&0.141$\pm$0.015	        &0.147$\pm$0.008$\bullet$	&0.167$\pm$0.011$\bullet$
  &0.288$\pm$0.019$\bullet$	&0.332$\pm$0.050$\bullet$	\cr
    scene	&0.086$\pm$0.003	&0.117$\pm$0.005$\bullet$	&0.131$\pm$0.018$\bullet$	&0.086$\pm$0.006	        &0.112$\pm$0.005$\bullet$	&0.084$\pm$0.004
 &0.127$\pm$0.003$\bullet$	&0.151$\pm$0.005$\bullet$	\cr
    yeast	&0.173$\pm$0.003	&0.193$\pm$0.004$\bullet$	&0.195$\pm$0.004$\bullet$	&0.174$\pm$0.004	        &0.181$\pm$0.004$\bullet$	&0.182$\pm$0.003$\bullet$
 &0.198$\pm$0.003$\bullet$	&0.190$\pm$0.003$\bullet$	\cr
    slashdot	&0.118$\pm$0.003	&0.180$\pm$0.005$\bullet$	&0.153$\pm$0.005$\bullet$	&0.132$\pm$0.005$\bullet$	&0.141$\pm$0.005$\bullet$	&0.178$\pm$0.005$\bullet$
 &0.258$\pm$0.005$\bullet$	&0.123$\pm$0.005$\bullet$	\cr

    \midrule
    \multirow{2}{1.5cm}{Data set}&
    \multicolumn{8}{c}{One-error$\downarrow$}\cr
    \cmidrule(lr){2-9}
    &LEMLL&MLFE&ML$^2$&RELIAB&MSVR&ML-kNN&CLR&ECC\cr
    \midrule
    CAL500	&0.122$\pm$0.017	&0.140$\pm$0.031	        &0.258$\pm$0.146$\bullet$	&0.120$\pm$0.016          	&0.122$\pm$0.017           	&0.136$\pm$0.017 &0.331$\pm$0.117$\bullet$	&0.191$\pm$0.022$\bullet$	\cr
    medical	&0.140$\pm$0.010	&0.173$\pm$0.014$\bullet$	&0.245$\pm$0.043$\bullet$	&0.213$\pm$0.022$\bullet$	&0.163$\pm$0.012$\bullet$   &0.297$\pm$0.020$\bullet$
  &0.688$\pm$0.151$\bullet$	&0.182$\pm$0.019$\bullet$	\cr
    llog	&0.782$\pm$0.021	&0.792$\pm$0.013	        &0.800$\pm$0.014$\bullet$	&0.748$\pm$0.011$\circ$    	&0.808$\pm$0.012$\bullet$	&0.802$\pm$0.013$\bullet$
    &0.883$\pm$0.024$\bullet$	&0.785$\pm$0.009	\cr
    enron	&0.241$\pm$0.013	&0.392$\pm$0.016$\bullet$	&0.662$\pm$0.049$\bullet$	&0.311$\pm$0.013$\bullet$	&0.384$\pm$0.013$\bullet$	&0.328$\pm$0.013$\bullet$
    &0.376$\pm$0.017$\bullet$	&0.424$\pm$0.013$\bullet$	\cr
    msra	&0.051$\pm$0.019	&0.087$\pm$0.014$\bullet$	&0.088$\pm$0.020$\bullet$	&0.097$\pm$0.029$\bullet$	&0.080$\pm$0.017$\bullet$	&0.081$\pm$0.020$\bullet$
 &0.312$\pm$0.089$\bullet$	&0.420$\pm$0.110$\bullet$	\cr
    scene	&0.253$\pm$0.010	&0.316$\pm$0.015$\bullet$	&0.337$\pm$0.034$\bullet$	&0.270$\pm$0.017$\bullet$	&0.296$\pm$0.013$\bullet$	&0.246$\pm$0.009
    &0.371$\pm$0.008$\bullet$	&0.373$\pm$0.009$\bullet$	\cr
    yeast	&0.233$\pm$0.009	&0.286$\pm$0.013$\bullet$	&0.308$\pm$0.011$\bullet$	&0.241$\pm$0.011	        &0.241$\pm$0.009$\bullet$	&0.247$\pm$0.010$\bullet$
 &0.270$\pm$0.007$\bullet$	&0.256$\pm$0.008$\bullet$	\cr
    slashdot	&0.411$\pm$0.008	&0.515$\pm$0.014$\bullet$	&0.463$\pm$0.011$\bullet$	&0.557$\pm$0.010$\bullet$	&0.428$\pm$0.010$\bullet$	&0.670$\pm$0.017$\bullet$
 &0.978$\pm$0.003$\bullet$	&0.481$\pm$0.014$\bullet$	\cr

    \midrule
    \multirow{2}{1.5cm}{Data set}&
    \multicolumn{8}{c}{Coverage$\downarrow$}\cr
    \cmidrule(lr){2-9}
    &LEMLL&MLFE&ML$^2$&RELIAB&MSVR&ML-kNN&CLR&ECC\cr
    \midrule
    CAL500	&0.748$\pm$0.008	&0.748$\pm$0.008	        &0.768$\pm$0.016$\bullet$	&0.747$\pm$0.008	        &0.748$\pm$0.008	        &0.755$\pm$0.006$\bullet$
  &0.794$\pm$0.010$\bullet$	&0.788$\pm$0.008$\bullet$	\cr
    medical	&0.040$\pm$0.007	&0.051$\pm$0.006$\bullet$	&0.077$\pm$0.009$\bullet$	&0.052$\pm$0.009$\bullet$	&0.056$\pm$0.009$\bullet$	&0.076$\pm$0.010$\bullet$
   &0.143$\pm$0.032$\bullet$	&0.048$\pm$0.010	\cr
    llog	&0.150$\pm$0.009	&0.261$\pm$0.015$\bullet$	&0.299$\pm$0.013$\bullet$	&0.159$\pm$0.006$\bullet$	&0.298$\pm$0.010$\bullet$	&0.169$\pm$0.009$\bullet$
 &0.234$\pm$0.020$\bullet$	&0.192$\pm$0.011$\bullet$	\cr
    enron	&0.245$\pm$0.007	&0.452$\pm$0.013$\bullet$	&0.526$\pm$0.031$\bullet$	&0.241$\pm$0.005	        &0.447$\pm$0.011$\bullet$	&0.265$\pm$0.006$\bullet$
 &0.238$\pm$0.006$\circ$	    &0.300$\pm$0.010$\bullet$	\cr
    msra	&0.544$\pm$0.017	&0.581$\pm$0.015$\bullet$	&0.585$\pm$0.016$\bullet$	&0.543$\pm$0.021	        &0.566$\pm$0.015$\bullet$	&0.590$\pm$0.013$\bullet$
  &0.720$\pm$0.024$\bullet$	&0.743$\pm$0.034$\bullet$	\cr
    scene	&0.085$\pm$0.002	&0.111$\pm$0.004$\bullet$	&0.124$\pm$0.015$\bullet$	&0.103$\pm$0.006$\bullet$	&0.108$\pm$0.004$\bullet$	&0.084$\pm$0.003
    &0.144$\pm$0.003$\bullet$	&0.169$\pm$0.004$\bullet$	\cr
    yeast	&0.455$\pm$0.005	&0.479$\pm$0.006$\bullet$	&0.480$\pm$0.006$\bullet$	&0.451$\pm$0.006	        &0.471$\pm$0.006$\bullet$	&0.465$\pm$0.005$\bullet$
 &0.492$\pm$0.006$\bullet$	&0.476$\pm$0.004$\bullet$	\cr
    slashdot	&0.137$\pm$0.004	&0.200$\pm$0.006$\bullet$	&0.174$\pm$0.005$\bullet$	&0.148$\pm$0.005$\bullet$	&0.161$\pm$0.005$\bullet$	&0.191$\pm$0.005$\bullet$
 &0.271$\pm$0.005$\bullet$	&0.139$\pm$0.005	\cr

    \midrule
    \multirow{2}{1.5cm}{Data set}&
    \multicolumn{8}{c}{Average precision$\uparrow$}\cr
    \cmidrule(lr){2-9}
    &LEMLL&MLFE&ML$^2$&RELIAB&MSVR&ML-kNN&CLR&ECC\cr
    \midrule
    CAL500	&0.497$\pm$0.005	&0.488$\pm$0.005$\bullet$	&0.435$\pm$0.032$\bullet$	&0.496$\pm$0.005	        &0.497$\pm$0.005	        &0.479$\pm$0.006$\bullet$
 &0.395$\pm$0.045$\bullet$	&0.462$\pm$0.007$\bullet$	\cr
    medical	&0.891$\pm$0.011	&0.865$\pm$0.013$\bullet$	&0.795$\pm$0.033$\bullet$	&0.837$\pm$0.019$\bullet$	&0.869$\pm$0.011$\bullet$	&0.770$\pm$0.017$\bullet$
   &0.400$\pm$0.065$\bullet$	&0.860$\pm$0.016$\bullet$	\cr
    llog	&0.337$\pm$0.012	&0.300$\pm$0.010$\bullet$	&0.274$\pm$0.009$\bullet$	&0.390$\pm$0.010$\circ$	&0.274$\pm$0.008$\bullet$	    &0.304$\pm$0.009$\bullet$
   &0.209$\pm$0.020$\bullet$	&0.342$\pm$0.009	\cr
    enron	&0.678$\pm$0.006	&0.539$\pm$0.011$\bullet$	&0.368$\pm$0.017$\bullet$	&0.661$\pm$0.008$\bullet$	&0.555$\pm$0.008$\bullet$	&0.609$\pm$0.010$\bullet$
    &0.610$\pm$0.008$\bullet$	&0.559$\pm$0.008$\bullet$	\cr
    msra	&0.816$\pm$0.014	&0.783$\pm$0.011$\bullet$	&0.774$\pm$0.010$\bullet$	&0.800$\pm$0.021	        &0.800$\pm$0.012$\bullet$	&0.775$\pm$0.016$\bullet$
  &0.624$\pm$0.023$\bullet$	&0.567$\pm$0.050$\bullet$	\cr
    scene	&0.850$\pm$0.005	&0.807$\pm$0.008$\bullet$	&0.792$\pm$0.022$\bullet$	&0.842$\pm$0.010$\bullet$	&0.818$\pm$0.008$\bullet$	&0.853$\pm$0.005
    &0.778$\pm$0.004$\bullet$	&0.766$\pm$0.006$\bullet$	\cr
    yeast	&0.754$\pm$0.005	&0.727$\pm$0.005$\bullet$	&0.720$\pm$0.005$\bullet$	&0.753$\pm$0.006	        &0.752$\pm$0.005	        &0.744$\pm$0.005$\bullet$
  &0.730$\pm$0.003$\bullet$	&0.741$\pm$0.004$\bullet$	\cr
    slashdot	&0.683$\pm$0.007	&0.594$\pm$0.010$\bullet$	&0.636$\pm$0.007$\bullet$	&0.579$\pm$0.009$\bullet$	&0.663$\pm$0.006$\bullet$	&0.480$\pm$0.012$\bullet$
 &0.251$\pm$0.007$\bullet$	&0.628$\pm$0.009$\bullet$	\cr

    \midrule
    \midrule
    \end{tabular}
\end{table*}

\renewcommand{\multirowsetup}{\centering}
\begin{table*}[!htb]
\centering
  \fontsize{7}{7}\selectfont
  \caption{Predictive performance (mean $\pm$ std. deviation). $\bullet$($\circ$) indicates LEMLL is significantly better (worse) than the corresponding method on the criterion based on paired \emph{t}-test at 95$\%$ significance level. $\downarrow$ ($\uparrow$) implies the smaller (larger), the better.}
    \begin{tabular}{ccccccccc}
    \midrule
    \midrule
    \multirow{2}{1.5cm}{Data set}&
    \multicolumn{8}{c}{Hamming loss$\downarrow$}\cr
    \cmidrule(lr){2-9}
    &LEMLL&MLFE&ML$^2$&RELIAB&MSVR&ML-kNN&CLR&ECC\cr
    \midrule
    corel5k	    &0.009$\pm$0.000	&0.013$\pm$0.000$\bullet$  &0.019$\pm$0.000$\bullet$    &0.024$\pm$0.000$\bullet$	&0.010$\pm$0.000$\bullet$	&0.009$\pm$0.000
 &0.011$\pm$0.000$\bullet$   &0.010$\pm$0.000$\bullet$	\cr
    rcv1-s1	&0.027$\pm$0.000	&0.035$\pm$0.003$\bullet$  &0.045$\pm$0.003$\bullet$	&0.031$\pm$0.000$\bullet$	&0.027$\pm$0.000           	&0.027$\pm$0.000
   &0.035$\pm$0.001$\bullet$	&0.029$\pm$0.000$\bullet$	\cr
    rcv1-s2	&0.024$\pm$0.000	&0.033$\pm$0.003$\bullet$  &0.051$\pm$0.004$\bullet$	&0.028$\pm$0.000$\bullet$	&0.024$\pm$0.000           	&0.025$\pm$0.000$\bullet$
   &0.034$\pm$0.001$\bullet$	&0.028$\pm$0.001$\bullet$	\cr
    bibtex  	&0.013$\pm$0.000	&0.015$\pm$0.000$\bullet$  &0.026$\pm$0.002$\bullet$	&0.014$\pm$0.000$\bullet$	&0.013$\pm$0.000          	&0.014$\pm$0.000$\bullet$
   &0.050$\pm$0.002$\bullet$	&0.021$\pm$0.001$\bullet$	\cr

    corel16k-s1	&0.019$\pm$0.000	&0.020$\pm$0.000$\bullet$  &0.019$\pm$0.000            &0.022$\pm$0.000$\bullet$	&0.019$\pm$0.000              &0.019$\pm$0.000
   &0.020$\pm$0.000$\bullet$	&0.019$\pm$0.000	\cr
    corel16k-s2	&0.017$\pm$0.000	&0.019$\pm$0.000$\bullet$  &0.018$\pm$0.000$\bullet$   &0.021$\pm$0.000$\bullet$	&0.018$\pm$0.000$\bullet$     &0.018$\pm$0.000$\bullet$   &0.019$\pm$0.000$\bullet$	&0.018$\pm$0.000	\cr

    tmc2007     &0.063$\pm$0.000    &0.066$\pm$0.000$\bullet$  &0.063$\pm$0.000            &0.066$\pm$0.001$\bullet$    &0.063$\pm$0.000              &0.075$\pm$0.000$\bullet$
 &0.072$\pm$0.001$\bullet$ &0.067$\pm$0.000$\bullet$    \cr
    \midrule
    \multirow{2}{1.5cm}{Data set}&
    \multicolumn{8}{c}{Ranking loss$\downarrow$}\cr
    \cmidrule(lr){2-9}
    &LEMLL&MLFE&ML$^2$&RELIAB&MSVR&ML-kNN&CLR&ECC\cr
    \midrule
    corel5k	    &0.134$\pm$0.002    &0.248$\pm$0.004$\bullet$	&0.462$\pm$0.015$\bullet$	&0.145$\pm$0.002$\bullet$	&0.236$\pm$0.003$\bullet$	&0.137$\pm$0.002$\bullet$
   &0.151$\pm$0.008$\bullet$	&0.181$\pm$0.002$\bullet$	\cr
    rcv1-s1	&0.044$\pm$0.001    &0.109$\pm$0.002$\bullet$	&0.129$\pm$0.008$\bullet$	&0.065$\pm$0.001$\bullet$	&0.090$\pm$0.002$\bullet$	&0.088$\pm$0.002$\bullet$
   &0.046$\pm$0.001$\bullet$	&0.083$\pm$0.002$\bullet$	\cr
    rcv1-s2	&0.045$\pm$0.001    &0.110$\pm$0.002$\bullet$	&0.145$\pm$0.011$\bullet$	&0.062$\pm$0.003$\bullet$	&0.090$\pm$0.003$\bullet$	&0.098$\pm$0.003$\bullet$
   &0.050$\pm$0.001$\bullet$   &0.087$\pm$0.002$\bullet$	\cr
    bibtex  	&0.074$\pm$0.002    &0.126$\pm$0.002$\bullet$   &0.138$\pm$0.005$\bullet$	&0.060$\pm$0.002$\circ$  	&0.132$\pm$0.002$\bullet$	&0.226$\pm$0.006$\bullet$
    &0.079$\pm$0.001$\bullet$	&0.145$\pm$0.002$\bullet$	\cr

    corel16k-s1	&0.150$\pm$0.001	&0.195$\pm$0.001$\bullet$  &0.198$\pm$0.003$\bullet$   &0.166$\pm$0.002$\bullet$	&0.196$\pm$0.002$\bullet$	&0.175$\pm$0.001$\bullet$
   &0.163$\pm$0.001$\bullet$	&0.227$\pm$0.002$\bullet$	\cr
    corel16k-s2	&0.166$\pm$0.001	&0.194$\pm$0.001$\bullet$  &0.197$\pm$0.001$\bullet$    &0.161$\pm$0.001	&0.194$\pm$0.001$\bullet$	&0.170$\pm$0.002
   &0.155$\pm$0.001$\circ$	&0.220$\pm$0.002$\bullet$	\cr

    tmc2007     &0.054$\pm$0.001    &0.054$\pm$0.001  &0.055$\pm$0.001$\bullet$    &0.048$\pm$0.001$\circ$     &0.055$\pm$0.001$\bullet$   &0.098$\pm$0.002$\bullet$
    &0.063$\pm$0.001$\bullet$   &0.067$\pm$0.001$\bullet$  \cr
    \midrule
    \multirow{2}{1.5cm}{Data set}&
    \multicolumn{8}{c}{One-error$\downarrow$}\cr
    \cmidrule(lr){2-9}
    &LEMLL&MLFE&ML$^2$&RELIAB&MSVR&ML-kNN&CLR&ECC\cr
    \midrule
    corel5k 	&0.662$\pm$0.009	&0.734$\pm$0.003$\bullet$      &0.828$\pm$0.011$\bullet$	&0.755$\pm$0.005$\bullet$	&0.672$\pm$0.005$\bullet$	&0.744$\pm$0.012$\bullet$
    &0.749$\pm$0.007$\bullet$	&0.747$\pm$0.006$\bullet$	\cr
    rcv1-s1	&0.438$\pm$0.004	&0.474$\pm$0.007$\bullet$      &0.571$\pm$0.026$\bullet$	&0.504$\pm$0.007$\bullet$	&0.453$\pm$0.007$\bullet$	&0.510$\pm$0.007$\bullet$
 &0.484$\pm$0.008$\bullet$	&0.527$\pm$0.010$\bullet$	\cr
    rcv1-s2	&0.436$\pm$0.011	&0.471$\pm$0.009$\bullet$      &0.588$\pm$0.026$\bullet$	&0.476$\pm$0.009$\bullet$	&0.449$\pm$0.008$\bullet$	&0.524$\pm$0.012$\bullet$
 &0.463$\pm$0.007$\bullet$	&0.520$\pm$0.009$\bullet$	\cr
    bibtex  	&0.395$\pm$0.005	&0.407$\pm$0.005$\bullet$      &0.581$\pm$0.022$\bullet$	&0.411$\pm$0.008$\bullet$	&0.396$\pm$0.005          	&0.610$\pm$0.005$\bullet$
   &0.510$\pm$0.006$\bullet$	&0.519$\pm$0.007$\bullet$	\cr

    corel16k-s1	&0.655$\pm$0.004	&0.689$\pm$0.004$\bullet$      &0.659$\pm$0.004             &0.715$\pm$0.006$\bullet$	&0.656$\pm$0.005	        &0.747$\pm$0.006$\bullet$
   &0.762$\pm$0.006$\bullet$	&0.787$\pm$0.005$\bullet$	\cr
    corel16k-s2	&0.644$\pm$0.003	&0.680$\pm$0.004$\bullet$      &0.652$\pm$0.005          &0.714$\pm$0.006$\bullet$	&0.650$\pm$0.004            &0.751$\pm$0.004$\bullet$
    &0.758$\pm$0.005$\bullet$	&0.778$\pm$0.007$\bullet$	\cr

    tmc2007     &0.238$\pm$0.005    &0.235$\pm$0.004               &0.238$\pm$0.005             &0.238$\pm$0.007            &0.238$\pm$0.005            &0.320$\pm$0.004$\bullet$
   &0.271$\pm$0.005$\bullet$        &0.239$\pm$0.003\cr
    \midrule
    \multirow{2}{1.5cm}{Data set}&
    \multicolumn{8}{c}{Coverage$\downarrow$}\cr
    \cmidrule(lr){2-9}
    &LEMLL&MLFE&ML$^2$&RELIAB&MSVR&ML-kNN&CLR&ECC\cr
    \midrule
    corel5k 	&0.322$\pm$0.004	&0.523$\pm$0.006$\bullet$  &0.765$\pm$0.014$\bullet$	&0.328$\pm$0.005$\bullet$	&0.518$\pm$0.006$\bullet$	&0.312$\pm$0.004$\circ$
    &0.314$\pm$0.013	        &0.416$\pm$0.005$\bullet$	\cr
    rcv1-s1	&0.110$\pm$0.001	&0.229$\pm$0.005$\bullet$  &0.256$\pm$0.012$\bullet$	&0.147$\pm$0.003$\bullet$	&0.201$\pm$0.003$\bullet$	&0.188$\pm$0.003$\bullet$
 &0.112$\pm$0.001$\bullet$	    &0.179$\pm$0.003$\bullet$	\cr
    rcv1-s2	&0.108$\pm$0.003	&0.223$\pm$0.002$\bullet$  &0.271$\pm$0.015$\bullet$	&0.134$\pm$0.005$\bullet$	&0.194$\pm$0.004$\bullet$	&0.201$\pm$0.004$\bullet$
 &0.115$\pm$0.002$\bullet$	    &0.181$\pm$0.004$\bullet$	\cr
    bibtex  	&0.140$\pm$0.003	&0.233$\pm$0.004$\bullet$  &0.233$\pm$0.007$\bullet$	&0.110$\pm$0.002$\circ$ 	&0.243$\pm$0.004$\bullet$	&0.365$\pm$0.009$\bullet$
 &0.135$\pm$0.001$\circ$	    &0.253$\pm$0.003$\bullet$	\cr

    corel16k-s1	&0.298$\pm$0.002	&0.378$\pm$0.003$\bullet$  &0.386$\pm$0.005$\bullet$    &0.324$\pm$0.003$\bullet$	&0.382$\pm$0.004$\bullet$	&0.339$\pm$0.002$\bullet$
  &0.303$\pm$0.002$\bullet$	&0.416$\pm$0.004$\bullet$	\cr
    corel16k-s2	&0.331$\pm$0.002	&0.377$\pm$0.003$\bullet$  &0.386$\pm$0.002$\bullet$    &0.317$\pm$0.003$\circ$	    &0.382$\pm$0.002$\bullet$	&0.333$\pm$0.003
    &0.288$\pm$0.002$\circ$     &0.406$\pm$0.003$\bullet$	\cr
    tmc2007     &0.135$\pm$0.001    &0.134$\pm$0.001           &0.137$\pm$0.001             &0.122$\pm$0.001$\circ$     &0.137$\pm$0.001            &0.195$\pm$0.002$\bullet$
   &0.145$\pm$0.001$\bullet$   &0.157$\pm$0.001$\bullet$   \cr
    \midrule
    \multirow{2}{1.5cm}{Data set}&
    \multicolumn{8}{c}{Average precision$\uparrow$}\cr
    \cmidrule(lr){2-9}
    &LEMLL&MLFE&ML$^2$&RELIAB&MSVR&ML-kNN&CLR&ECC\cr
    \midrule
    corel5k 	&0.293$\pm$0.003	&0.224$\pm$0.003$\bullet$     &0.144$\pm$0.003$\bullet$    &0.241$\pm$0.002$\bullet$	&0.268$\pm$0.002$\bullet$	&0.240$\pm$0.005$\bullet$
  &0.221$\pm$0.006$\bullet$	&0.234$\pm$0.004$\bullet$	\cr
    rcv1-s1	&0.593$\pm$0.002	&0.526$\pm$0.006$\bullet$     &0.440$\pm$0.017$\bullet$    &0.536$\pm$0.006$\bullet$	&0.559$\pm$0.005$\bullet$	&0.513$\pm$0.003$\bullet$
   &0.580$\pm$0.004$\bullet$	&0.498$\pm$0.006$\bullet$	\cr
    rcv1-s2	&0.605$\pm$0.005	&0.549$\pm$0.005$\bullet$     &0.445$\pm$0.020$\bullet$    &0.563$\pm$0.008$\bullet$	&0.580$\pm$0.006$\bullet$	&0.515$\pm$0.008$\bullet$
   &0.596$\pm$0.003$\bullet$   &0.516$\pm$0.005$\bullet$	\cr
    bibtex  	&0.568$\pm$0.004	&0.524$\pm$0.003$\bullet$     &0.396$\pm$0.013$\bullet$    &0.564$\pm$0.004        	    &0.529$\pm$0.004$\bullet$	&0.327$\pm$0.006$\bullet$
   &0.473$\pm$0.003$\bullet$	&0.412$\pm$0.005$\bullet$	\cr

    corel16k-s1	&0.337$\pm$0.002	&0.310$\pm$0.003$\bullet$     &0.324$\pm$0.003$\bullet$    &0.299$\pm$0.003$\bullet$	&0.325$\pm$0.003$\bullet$	&0.279$\pm$0.002$\bullet$
   &0.260$\pm$0.003$\bullet$	&0.231$\pm$0.003$\bullet$	\cr
    corel16k-s2	&0.332$\pm$0.002	&0.307$\pm$0.003$\bullet$     &0.321$\pm$0.003$\bullet$    &0.296$\pm$0.002$\bullet$	&0.322$\pm$0.002$\bullet$	&0.272$\pm$0.002$\bullet$
   &0.260$\pm$0.003$\bullet$	&0.232$\pm$0.003$\bullet$	\cr

    tmc2007     &0.800$\pm$0.003    &0.801$\pm$0.002              &0.800$\pm$0.003             &0.800$\pm$0.004             &0.800$\pm$0.003            &0.712$\pm$0.003$\bullet$
   &0.768$\pm$0.002$\bullet$    &0.783$\pm$0.002$\bullet$\cr
    \midrule
    \midrule
    \end{tabular}
\end{table*}

\begin{table}[!htb]
\caption{The average ranks of the 8 algorithms on the 5 measures.}
\fontsize{7}{7}\selectfont
\center
\begin{tabular}{c c c c c c}
\midrule\midrule
Algorithm & HL & RL & OE & CO & AP \\
\midrule
LEMLL   &1.500	&1.567	&1.400	&2.000	&1.400\\
MLFE    &5.567	&5.400	&4.067	&5.233	&4.600\\
ML$^2$  &6.033	&6.867	&5.900	&6.600	&6.067\\
RELIAB  &5.567	&2.233	&4.067	&2.667	&3.400\\
MSVR    &2.833	&4.833	&2.833	&5.033	&3.067\\
ML-kNN  &3.300	&4.833	&5.333	&4.667	&5.467\\
CLR     &6.667	&4.800	&6.333	&4.600	&6.133\\
ECC     &4.533	&5.467	&6.067	&5.600	&5.867\\
\midrule\midrule
\end{tabular}
\end{table}

\begin{table}[!htb]
\caption{Characteristics of the 15 data sets used in the experiments.}
\fontsize{7}{7}\selectfont
\center
\begin{tabular}{c c c c c c c c c}
\midrule\midrule
Data set (\emph{abbr.}) & $|S|$ & $dim(S)$ & $L(S)$ \\
\midrule

SJAFFE (SJA) &213 &243 &6\\
Natural Scene (NS) &2,000 &294 &9\\
Yeast-spoem (spoem) &2,465 &24 &2\\
Yeast-spo5 (spo5) &2,465 &24 &3\\
Yeast-dtt (dtt) &2,465 &24 &4\\
Yeast-cold (cold) &2,465 &24 &4\\
Yeast-spo (spo) &2,465 &24 &6\\
Yeast-heat (heat) &2,465 &24 &6\\
Yeast-diau (diau) &2,465 &24 &7\\
Yeast-elu (elu) &2,465 &24 &14\\
Yeast-cdc (cdc) &2,465 &24 &15\\
Yeast-alpha (alpha) &2,465 &24 &18\\
SBU\_3DFE (3DFE) &2,500 &243 &6\\
Movie (Mov) &7,755 &1,869 &5\\
Human Gene (HG) &30,542 &36 &68\\
\midrule\midrule
\end{tabular}
\end{table}

\subsubsection{Experimental Results}
The detailed experimental results of each comparing algorithm on the 15 data sets are presented in Table II and Table III. The average ranks of the eight algorithms on the five measures are given in Table IV. On each data set, 50\% examples are randomly sampled without replacement to form the training set, and the rest 50\% examples are used to form the test set. The sampling process is repeated for ten times. The mean metric value and the standard deviation across ten training/testing trials are recorded for comparative studies.

\renewcommand{\multirowsetup}{\centering}
\begin{table*}[!htb]
\centering
  \fontsize{6}{6}\selectfont
  \caption{Reconstruction performance (value(rank)) measured by Chebyshev with threshold $\rho$ varying from 0.1 to 0.5 with step size of 0.1}
    \begin{tabular}{p{0.64cm}<{\centering} p{0.64cm}<{\centering} p{0.64cm}<{\centering} p{0.64cm}<{\centering} p{0.64cm}<{\centering} p{0.64cm}<{\centering} p{0.64cm}<{\centering} p{0.64cm}<{\centering} p{0.64cm}<{\centering} p{0.64cm}<{\centering} p{0.64cm}<{\centering} p{0.64cm}<{\centering} p{0.64cm}<{\centering} p{0.64cm}<{\centering} p{0.64cm}<{\centering} p{0.64cm}<{\centering} p{0.6cm}<{\centering}}
    \midrule
    \midrule
    \multirow{2}{0.7cm}{Algorithm}&
    \multicolumn{15}{c}{$\rho=0.1$}&\multirow{2}{0.7cm}{Avg. Rank}\cr
    \cmidrule(lr){2-16}
    &SJA&NS&spoem&spo5&dtt&cold&heat&spo&diau&elu&cdc&alpha&3DFE&Mov&HG \cr
    \midrule

LEMLL	&\bf{0.077(1)}	&0.328(3)      	&\bf{0.063(1)}	&\bf{0.084(1)}	&\bf{0.069(1)}	&\bf{0.069(1)}	&\bf{0.054(1)}	&\bf{0.065(1)}	&\bf{0.052(1)}	&\bf{0.026(1)}	&\bf{0.026(1)}	&\bf{0.023(1)}	&\bf{0.087(1)}	&\bf{0.117(1)}	&0.048(2)	&1.200\cr
MLFE	&0.093(2)      	&\bf{0.313(1)}	&0.149(4)	&0.176(3)	&0.203(3)	&0.189(2)	&0.161(2)	&0.148(2)	&0.155(2)	&0.081(3)	&0.078(2)	&0.071(2)	&0.088(2)	&0.143(2)	&\bf{0.044(1)}	&2.200\cr
ML$^2$	&0.122(3)    	&0.316(2)      	&0.165(5)	&0.220(4)	&0.265(5)	&0.265(4)	&0.273(5)	&0.269(3)	&0.312(5)	&0.191(5)	&0.195(5)	&0.194(5)	&0.123(3)	&0.236(3)	&0.055(3)	&4.000\cr
RELIAB	&0.159(4)    	&0.427(6)      	&0.081(2)	&0.173(2)	&0.208(4)	&0.210(3)	&0.234(4)	&0.278(4)	&0.285(4)	&0.135(4)	&0.148(4)	&0.146(4)	&0.164(4)	&0.379(5)	&0.195(5)	&3.933\cr
FCM 	&0.163(5)      	&0.393(5)      	&0.089(3)	&0.252(5)	&0.173(2)	&0.295(5)	&0.164(3)	&0.350(5)	&0.175(3)	&0.080(2)	&0.084(3)	&0.086(3)	&0.203(5)	&0.364(4)	&0.061(4)	&3.800\cr
KM  	&0.718(6)      	&0.348(4)      	&0.411(6)	&0.586(6)	&0.719(6)	&0.703(6)	&0.796(6)	&0.779(6)	&0.824(6)	&0.461(6)	&0.460(6)	&0.469(6)	&0.695(6)	&0.624(6)	&0.365(6)	&5.867\cr

    \midrule
    \multirow{2}{0.7cm}{Algorithm}&
    \multicolumn{15}{c}{$\rho=0.2$}&\multirow{2}{0.7cm}{Avg. Rank}\cr
    \cmidrule(lr){2-16}
    &SJA&NS&spoem&spo5&dtt&cold&heat&spo&diau&elu&cdc&alpha&3DFE&Mov&HG \cr
    \midrule
LEMLL	&\bf{0.077(1)}	&0.328(3)	&\bf{0.063(1)}	&\bf{0.084(1)}	&\bf{0.069(1)}	&\bf{0.069(1)}	&\bf{0.049(1)}	&\bf{0.061(1)}	&\bf{0.052(1)}	&\bf{0.025(1)}	&\bf{0.026(1)}	&\bf{0.022(1)}	&0.087(2)	&\bf{0.117(1)}	&0.048(3)	&1.333\cr
MLFE	&0.092(2)	&0.312(2)	&0.149(4)	&0.176(3)	&0.203(3)	&0.189(2)	&0.127(3)	&0.123(2)	&0.106(2)	&0.067(3)	&0.065(2)	&0.053(3)	&\bf{0.085(1)}	&0.143(2)	&\bf{0.042(1)}	&2.333\cr
ML$^2$	&0.117(3)	&\bf{0.308(1)}	&0.165(5)	&0.220(4)	&0.265(5)	&0.265(4)	&0.198(5)	&0.210(4)	&0.173(5)	&0.133(5)	&0.138(5)	&0.116(5)	&0.116(3)	&0.236(3)	&0.046(2)	&3.933\cr
RELIAB	&0.153(4)	&0.418(6)	&0.081(2)	&0.173(2)	&0.208(4)	&0.210(3)	&0.138(4)	&0.200(3)	&0.155(4)	&0.082(4)	&0.095(4)	&0.072(4)	&0.155(4)	&0.379(5)	&0.088(5)	&3.867\cr
FCM	&0.161(5)	&0.394(5)	&0.089(3)	&0.252(5)	&0.173(2)	&0.295(5)	&0.115(2)	&0.272(5)	&0.135(3)	&0.050(2)	&0.07(3)	&0.050(2)	&0.199(5)	&0.364(4)	&0.056(4)	&3.667\cr
KM	&0.706(6)	&0.348(4)	&0.411(6)	&0.586(6)	&0.719(6)	&0.703(6)	&0.565(6)	&0.621(6)	&0.392(6)	&0.278(6)	&0.281(6)	&0.223(6)	&0.666(6)	&0.624(6)	&0.179(6)	&5.867\cr

    \midrule
    \multirow{2}{0.7cm}{Algorithm}&
    \multicolumn{15}{c}{$\rho=0.3$}&\multirow{2}{0.7cm}{Avg. Rank}\cr
    \cmidrule(lr){2-16}
    &SJA&NS&spoem&spo5&dtt&cold&heat&spo&diau&elu&cdc&alpha&3DFE&Mov&HG \cr
    \midrule

LEMLL	&\bf{0.073(1)}	&0.323(4)	&\bf{0.063(1)}	&\bf{0.084(1)}	&\bf{0.055(1)}	&\bf{0.061(1)}	&\bf{0.046(1)}	&\bf{0.052(1)}	&\bf{0.052(1)}	&\bf{0.024(1)}	&\bf{0.023(1)}	&\bf{0.020(1)}	&0.088(2)	&\bf{0.116(1)}	&0.048(3)	&1.400\cr
MLFE	&0.080(2)	&0.308(3)	&0.149(4)	&0.176(3)	&0.122(4)	&0.130(2.5)	&0.108(4)	&0.101(2)	&0.104(2)	&0.055(3)	&0.050(2)	&0.044(2)	&\bf{0.078(1)}	&0.140(2)	&\bf{0.042(1)}	&2.500\cr
ML$^2$	&0.089(3)	&0.298(2)	&0.165(5)	&0.220(4)	&0.151(5)	&0.168(4)	&0.151(5)	&0.147(4)	&0.166(5)	&0.103(5)	&0.098(5)	&0.089(5)	&0.092(3)	&0.223(3)	&0.043(2)	&4.000\cr
RELIAB	&0.096(4)	&\bf{0.284(1)}	&0.081(2)	&0.173(2)	&0.089(2)	&0.130(2.5)	&0.093(2)	&0.111(3)	&0.149(4)	&0.058(4)	&0.058(4)	&0.052(4)	&0.108(4)	&0.357(5)	&0.055(5)	&3.233\cr
FCM	&0.183(5)	&0.382(6)	&0.089(3)	&0.252(5)	&0.102(3)	&0.223(5)	&0.106(3)	&0.167(5)	&0.133(3)	&0.046(2)	&0.057(3)	&0.047(3)	&0.182(5)	&0.351(4)	&0.054(4)	&3.933\cr
KM	&0.408(6)	&0.346(5)	&0.411(6)	&0.586(6)	&0.316(6)	&0.414(6)	&0.342(6)	&0.339(6)	&0.352(6)	&0.187(6)	&0.165(6)	&0.148(6)	&0.455(6)	&0.585(6)	&0.105(6)	&5.933\cr

    \midrule
    \multirow{2}{0.7cm}{Algorithm}&
    \multicolumn{15}{c}{$\rho=0.4$}&\multirow{2}{0.7cm}{Avg. Rank}\cr
    \cmidrule(lr){2-16}
    &SJA&NS&spoem&spo5&dtt&cold&heat&spo&diau&elu&cdc&alpha&3DFE&Mov&HG \cr
    \midrule

LEMLL	&\bf{0.078(1)}	&0.320(4)	&\bf{0.063(1)}	&\bf{0.068(1)}	&\bf{0.053(1)}	&\bf{0.056(1)}	&\bf{0.041(1)}	&\bf{0.049(1)}	&\bf{0.051(1)}	&\bf{0.022(1)}	&\bf{0.021(1)}	&\bf{0.019(1)}	&\bf{0.095(1)}	&\bf{0.101(1)}	&0.049(4)	&1.400\cr
MLFE	&0.093(2.5)	&0.306(3)	&0.149(4)	&0.138(3)	&0.114(4)	&0.111(3)	&0.084(4)	&0.087(3)	&0.077(2)	&0.042(3)	&0.040(2.5)	&0.034(2)	&0.098(2)	&0.106(2)	&0.043(2.5)	&2.833\cr
ML$^2$	&0.099(4)	&0.294(2)	&0.165(5)	&0.162(5)	&0.139(5)	&0.136(4)	&0.117(5)	&0.121(5)	&0.119(4)	&0.077(5)	&0.075(5)	&0.067(5)	&0.104(4)	&0.142(3)	&0.043(2.5)	&4.233\cr
RELIAB	&0.093(2.5)	&\bf{0.265(1)}	&0.081(2)	&0.089(2)	&0.084(2)	&0.102(2)	&0.062(2)	&0.082(2)	&0.116(3)	&0.040(2)	&0.040(2.5)	&0.041(3)	&0.103(3)	&0.203(4)	&\bf{0.042(1)}	&2.267\cr
FCM	&0.164(5)	&0.373(6)	&0.089(3)	&0.143(4)	&0.100(3)	&0.191(5)	&0.079(3)	&0.117(4)	&0.134(5)	&0.046(4)	&0.046(4)	&0.045(4)	&0.151(5)	&0.226(5)	&0.054(5)	&4.333\cr
KM	&0.308(6)	&0.346(5)	&0.411(6)	&0.420(6)	&0.257(6)	&0.254(6)	&0.217(6)	&0.239(6)	&0.195(6)	&0.113(6)	&0.109(6)	&0.095(6)	&0.337(6)	&0.321(6)	&0.067(6)	&5.933\cr

    \midrule
    \multirow{2}{0.7cm}{Algorithm}&
    \multicolumn{15}{c}{$\rho=0.5$}&\multirow{2}{0.7cm}{Avg. Rank}\cr
    \cmidrule(lr){2-16}
    &SJA&NS&spoem&spo5&dtt&cold&heat&spo&diau&elu&cdc&alpha&3DFE&Mov&HG \cr
    \midrule

LEMLL	&\bf{0.083(1)}	&0.318(4)	&\bf{0.063(1)}	&\bf{0.072(1)}	&\bf{0.053(1)}	&\bf{0.056(1)}	&\bf{0.041(1)}	&\bf{0.047(1)}	&\bf{0.044(1)}	&\bf{0.020(1)}	&\bf{0.019(1)}	&\bf{0.018(1)}	&0.107(3)	&0.105(2)	&0.047(3)	&1.533\cr
MLFE	&0.087(3)	&0.305(3)	&0.148(4)	&0.115(3)	&0.114(4)	&0.112(3)	&0.078(4)	&0.078(3)	&0.070(2)	&0.035(3)	&0.034(3)	&0.028(2)	&\bf{0.100(1)}	&\bf{0.102(1)}	&0.047(3)	&2.800\cr
ML$^2$	&0.093(4)	&0.291(2)	&0.163(5)	&0.130(4)	&0.139(5)	&0.136(4)	&0.106(5)	&0.103(5)	&0.100(4)	&0.063(5)	&0.062(5)	&0.052(5)	&0.105(2)	&0.129(3)	&0.047(3)	&4.067\cr
RELIAB	&0.086(2)	&\bf{0.257(1)}	&0.079(2)	&0.084(2)	&0.084(2)	&0.101(2)	&0.056(2)	&0.061(2)	&0.086(3)	&0.034(2)	&0.031(2)	&0.035(3)	&0.111(4)	&0.168(4)	&\bf{0.039(1)}	&2.267\cr
FCM	&0.157(5)	&0.369(6)	&0.089(3)	&0.154(5)	&0.100(3)	&0.189(5)	&0.070(3)	&0.088(4)	&0.124(5)	&0.045(4)	&0.042(4)	&0.042(4)	&0.158(5)	&0.207(5)	&0.055(5)	&4.400\cr
KM	&0.213(6)	&0.346(5)	&0.408(6)	&0.276(6)	&0.257(6)	&0.252(6)	&0.175(6)	&0.175(6)	&0.152(6)	&0.078(6)	&0.076(6)	&0.063(6)	&0.238(6)	&0.234(6)	&0.058(6)	&5.933\cr

    \midrule
    \midrule
    \end{tabular}
\end{table*}

\renewcommand{\multirowsetup}{\centering}
\begin{table*}[!htb]
\centering
  \fontsize{6}{6}\selectfont
  \caption{Reconstruction performance (value(rank)) measured by K-L with threshold $\rho$ varying from 0.1 to 0.5 with step size of 0.1}
    \begin{tabular}{p{0.64cm}<{\centering} p{0.64cm}<{\centering} p{0.64cm}<{\centering} p{0.64cm}<{\centering} p{0.64cm}<{\centering} p{0.64cm}<{\centering} p{0.64cm}<{\centering} p{0.64cm}<{\centering} p{0.64cm}<{\centering} p{0.64cm}<{\centering} p{0.64cm}<{\centering} p{0.64cm}<{\centering} p{0.64cm}<{\centering} p{0.64cm}<{\centering} p{0.64cm}<{\centering} p{0.64cm}<{\centering} p{0.6cm}<{\centering}}
    \midrule
    \midrule
    \multirow{2}{0.7cm}{Algorithm}&
    \multicolumn{15}{c}{$\rho=0.1$}&\multirow{2}{0.7cm}{Avg. Rank}\cr
    \cmidrule(lr){2-16}
    &SJA&NS&spoem&spo5&dtt&cold&heat&spo&diau&elu&cdc&alpha&3DFE&Mov&HG \cr
    \midrule

LEMLL	&\bf{0.038(1)}	&3.223(4)	&\bf{0.019(1)}	&\bf{0.028(1)}	&\bf{0.020(1)}	&\bf{0.022(1)}	&\bf{0.020(1)}	&\bf{0.030(1)}	&\bf{0.023(1)}	&\bf{0.014(1)}	&\bf{0.016(1)}	&\bf{0.013(1)}	&0.037(2)	&\bf{0.145(1)}	&0.213(2)	&1.333\cr
MLFE	&0.043(2)	&2.844(3)	&0.059(4)	&0.076(2)	&0.100(2)	&0.090(2)	&0.084(2)	&0.078(2)	&0.089(2)	&0.084(2)	&0.084(2)	&0.077(2)	&\bf{0.033(1)}	&0.149(2)	&\bf{0.200(1)}	&2.067\cr
ML$^2$	&0.073(3)	&2.588(2)	&0.078(5)	&0.136(4)	&0.211(5)	&0.224(4)	&0.317(5)	&0.328(4)	&0.467(5)	&0.620(5)	&0.677(5)	&0.772(5)	&0.081(3)	&0.279(3)	&0.430(3)	&4.067\cr
RELIAB	&0.163(4)	&\bf{1.361(1)}	&0.023(2)	&0.085(3)	&0.121(4)	&0.138(3)	&0.203(3)	&0.314(3)	&0.421(4)	&0.206(4)	&0.293(4)	&0.305(4)	&0.152(4)	&0.446(4)	&1.063(5)	&3.467\cr
FCM	&0.218(5)	&3.693(6)	&0.035(3)	&0.187(5)	&0.112(3)	&0.226(5)	&0.226(4)	&0.392(5)	&0.344(3)	&0.123(3)	&0.222(3)	&0.138(3)	&0.284(5)	&0.578(5)	&0.443(4)	&4.133\cr
KM	&1.287(6)	&3.518(5)	&0.536(6)	&0.891(6)	&1.271(6)	&1.218(6)	&1.595(6)	&1.523(6)	&1.742(6)	&1.825(6)	&1.880(6)	&2.054(6)	&1.218(6)	&0.990(6)	&2.099(6)	&5.933\cr

    \midrule
    \multirow{2}{0.7cm}{Algorithm}&
    \multicolumn{15}{c}{$\rho=0.2$}&\multirow{2}{0.7cm}{Avg. Rank}\cr
    \cmidrule(lr){2-16}
    &SJA&NS&spoem&spo5&dtt&cold&heat&spo&diau&elu&cdc&alpha&3DFE&Mov&HG \cr
    \midrule

LEMLL	&\bf{0.038(1)}	&3.220(4)	&\bf{0.019(1)}	&\bf{0.028(1)}	&\bf{0.020(1)}	&\bf{0.022(1)}	&\bf{0.019(1)}	&\bf{0.028(1)}	&\bf{0.027(1)}	&\bf{0.015(1)}	&\bf{0.017(1)}	&\bf{0.016(1)}	&0.037(2)	&\bf{0.145(1)}	&0.205(2)	&1.333\cr
MLFE	&0.043(2)	&2.840(3)	&0.059(4)	&0.076(2)	&0.100(2)	&0.090(2)	&0.084(2)	&0.077(2)	&0.092(2)	&0.095(3)	&0.093(2)	&0.096(3)	&\bf{0.033(1)}	&0.149(2)	&\bf{0.188(1)}	&2.200\cr
ML$^2$	&0.068(3)	&2.551(2)	&0.078(5)	&0.136(4)	&0.211(5)	&0.224(4)	&0.250(5)	&0.263(4)	&0.296(5)	&0.464(5)	&0.525(5)	&0.526(5)	&0.078(3)	&0.279(3)	&0.340(3)	&4.067\cr
RELIAB	&0.154(4)	&\bf{1.352(1)}	&0.023(2)	&0.085(3)	&0.121(4)	&0.138(3)	&0.105(3)	&0.191(3)	&0.244(3)	&0.131(4)	&0.189(4)	&0.164(4)	&0.145(4)	&0.446(4)	&0.710(5)	&3.400\cr
FCM	&0.208(5)	&3.694(6)	&0.035(3)	&0.187(5)	&0.112(3)	&0.226(5)	&0.110(4)	&0.269(5)	&0.249(4)	&0.088(2)	&0.157(3)	&0.079(2)	&0.278(5)	&0.578(5)	&0.351(4)	&4.067\cr
KM	&1.271(6)	&3.517(5)	&0.536(6)	&0.891(6)	&1.271(6)	&1.218(6)	&1.261(6)	&1.301(6)	&1.139(6)	&1.421(6)	&1.480(6)	&1.436(6)	&1.178(6)	&0.990(6)	&1.542(6)	&5.933\cr

    \midrule
    \multirow{2}{0.7cm}{Algorithm}&
    \multicolumn{15}{c}{$\rho=0.3$}&\multirow{2}{0.7cm}{Avg. Rank}\cr
    \cmidrule(lr){2-16}
    &SJA&NS&spoem&spo5&dtt&cold&heat&spo&diau&elu&cdc&alpha&3DFE&Mov&HG \cr
    \midrule

LEMLL	&\bf{0.030(1)}	&3.118(4)	&\bf{0.019(1)}	&\bf{0.028(1)}	&\bf{0.017(1)}	&\bf{0.019(1)}	&\bf{0.017(1)}	&\bf{0.023(1)}	&\bf{0.027(1)}	&\bf{0.016(1)}	&\bf{0.017(1)}	&\bf{0.017(1)}	&0.036(2)	&\bf{0.143(1)}	&0.196(2)	&1.333\cr
MLFE	&0.041(2)	&2.674(3)	&0.059(4)	&0.076(2)	&0.086(4)	&0.077(3)	&0.087(4)	&0.075(2)	&0.092(2)	&0.098(4)	&0.097(2)	&0.099(3)	&\bf{0.035(1)}	&0.146(2)	&\bf{0.178(1)}	&2.600\cr
ML$^2$	&0.053(3)	&2.323(2)	&0.078(5)	&0.136(4)	&0.134(5)	&0.140(5)	&0.203(5)	&0.192(5)	&0.287(5)	&0.374(5)	&0.377(5)	&0.413(5)	&0.059(3)	&0.267(3)	&0.279(3)	&4.200\cr
RELIAB	&0.061(4)	&\bf{1.196(1)}	&0.023(2)	&0.085(3)	&0.041(2)	&0.067(2)	&0.065(2)	&0.088(3)	&0.237(3)	&0.092(3)	&0.106(4)	&0.113(4)	&0.082(4)	&0.415(4)	&0.509(5)	&3.067\cr
FCM	&0.233(5)	&3.441(5)	&0.035(3)	&0.187(5)	&0.042(3)	&0.137(4)	&0.072(3)	&0.150(4)	&0.243(4)	&0.070(2)	&0.099(3)	&0.065(2)	&0.226(5)	&0.566(5)	&0.308(4)	&3.800\cr
KM	&0.883(6)	&3.475(6)	&0.536(6)	&0.891(6)	&0.701(6)	&0.806(6)	&0.959(6)	&0.925(6)	&1.086(6)	&1.141(6)	&1.093(6)	&1.141(6)	&0.907(6)	&0.936(6)	&1.189(6)	&6.000\cr

    \midrule
    \multirow{2}{0.7cm}{Algorithm}&
    \multicolumn{15}{c}{$\rho=0.4$}&\multirow{2}{0.7cm}{Avg. Rank}\cr
    \cmidrule(lr){2-16}
    &SJA&NS&spoem&spo5&dtt&cold&heat&spo&diau&elu&cdc&alpha&3DFE&Mov&HG \cr
    \midrule

LEMLL	&\bf{0.032(1)}	&3.056(4)	&\bf{0.019(1)}	&\bf{0.023(1)}	&\bf{0.016(1)}	&\bf{0.017(1)}	&\bf{0.015(1)}	&\bf{0.021(1)}	&\bf{0.027(1)}	&\bf{0.015(1)}	&\bf{0.016(1)}	&\bf{0.016(1)}	&\bf{0.038(1)}	&0.116(2)	&0.189(2)	&1.333\cr
MLFE	&0.047(2)	&2.578(3)	&0.059(4)	&0.055(3)	&0.084(4)	&0.073(3)	&0.075(4)	&0.068(3)	&0.081(2)	&0.092(4)	&0.093(4)	&0.096(4)	&0.046(2)	&\bf{0.103(1)}	&\bf{0.168(1)}	&2.933\cr
ML$^2$	&0.057(3.5)	&2.219(2)	&0.078(5)	&0.083(5)	&0.127(5)	&0.116(5)	&0.145(5)	&0.143(5)	&0.195(4)	&0.265(5)	&0.285(5)	&0.310(5)	&0.059(4)	&0.163(3)	&0.235(3)	&4.300\cr
RELIAB	&0.057(3.5)	&\bf{1.099(1)}	&0.023(2)	&0.031(2)	&0.037(2)	&0.046(2)	&0.034(2)	&0.052(2)	&0.142(3)	&0.058(2.5)	&0.065(2)	&0.076(3)	&0.057(3)	&0.251(4)	&0.377(5)	&2.600\cr
FCM	&0.186(5)	&3.284(5)	&0.035(3)	&0.074(4)	&0.041(3)	&0.101(4)	&0.036(3)	&0.104(4)	&0.200(5)	&0.058(2.5)	&0.067(3)	&0.054(2)	&0.142(5)	&0.375(5)	&0.280(4)	&3.833\cr
KM	&0.760(6)	&3.449(6)	&0.536(6)	&0.573(6)	&0.619(6)	&0.591(6)	&0.682(6)	&0.715(6)	&0.724(6)	&0.816(6)	&0.836(6)	&0.863(6)	&0.764(6)	&0.574(6)	&0.931(6)	&6.000\cr

    \midrule
    \multirow{2}{0.7cm}{Algorithm}&
    \multicolumn{15}{c}{$\rho=0.5$}&\multirow{2}{0.7cm}{Avg. Rank}\cr
    \cmidrule(lr){2-16}
    &SJA&NS&spoem&spo5&dtt&cold&heat&spo&diau&elu&cdc&alpha&3DFE&Mov&HG \cr
    \midrule

LEMLL	&\bf{0.032(1)}	&3.019(4)	&\bf{0.018(1)}	&\bf{0.021(1)}	&\bf{0.016(1)}	&\bf{0.016(1)}	&\bf{0.014(1)}	&\bf{0.018(1)}	&\bf{0.021(1)}	&\bf{0.013(1)}	&\bf{0.014(1)}	&\bf{0.014(1)}	&\bf{0.047(1)}	&0.108(2)	&0.183(2)	&1.333\cr
MLFE	&0.050(3)	&2.530(3)	&0.059(4)	&0.045(3)	&0.084(4)	&0.073(3)	&0.073(4)	&0.063(3)	&0.068(2)	&0.082(4)	&0.082(4)	&0.085(4)	&0.062(2)	&\bf{0.088(1)}	&\bf{0.158(1)}	&3.000\cr
ML$^2$	&0.058(4)	&2.181(2)	&0.078(5)	&0.061(4)	&0.127(5)	&0.115(5)	&0.131(5)	&0.117(5)	&0.140(4)	&0.204(5)	&0.217(5)	&0.228(5)	&0.072(4)	&0.137(3)	&0.201(3)	&4.267\cr
RELIAB	&0.041(2)	&\bf{1.083(1)}	&0.023(2)	&0.024(2)	&0.037(2)	&0.045(2)	&0.029(2)	&0.033(2)	&0.077(3)	&0.041(2)	&0.043(2)	&0.052(3)	&0.067(3)	&0.210(4)	&0.287(5)	&2.467\cr
FCM	&0.156(5)	&3.194(5)	&0.035(3)	&0.067(5)	&0.041(3)	&0.100(4)	&0.031(3)	&0.069(4)	&0.160(5)	&0.052(3)	&0.053(3)	&0.046(2)	&0.148(5)	&0.339(5)	&0.264(4)	&3.933\cr
KM	&0.561(6)	&3.432(6)	&0.532(6)	&0.334(6)	&0.619(6)	&0.588(6)	&0.589(6)	&0.564(6)	&0.540(6)	&0.621(6)	&0.635(6)	&0.636(6)	&0.606(6)	&0.454(6)	&0.729(6)	&6.000\cr

    \midrule
    \midrule
    \end{tabular}
\end{table*}

\renewcommand{\multirowsetup}{\centering}
\begin{table*}[!htb]
\centering
  \fontsize{6}{6}\selectfont
  \caption{Reconstruction performance (value(rank)) measured by Cosine with threshold $\rho$ varying from 0.1 to 0.5 with step size of 0.1}
    \begin{tabular}{p{0.64cm}<{\centering} p{0.64cm}<{\centering} p{0.64cm}<{\centering} p{0.64cm}<{\centering} p{0.64cm}<{\centering} p{0.64cm}<{\centering} p{0.64cm}<{\centering} p{0.64cm}<{\centering} p{0.64cm}<{\centering} p{0.64cm}<{\centering} p{0.64cm}<{\centering} p{0.64cm}<{\centering} p{0.64cm}<{\centering} p{0.64cm}<{\centering} p{0.64cm}<{\centering} p{0.64cm}<{\centering} p{0.6cm}<{\centering}}
    \midrule
    \midrule
    \multirow{2}{0.7cm}{Algorithm}&
    \multicolumn{15}{c}{$\rho=0.1$}&\multirow{2}{0.7cm}{Avg. Rank}\cr
    \cmidrule(lr){2-16}
    &SJA&NS&spoem&spo5&dtt&cold&heat&spo&diau&elu&cdc&alpha&3DFE&Mov&HG \cr
    \midrule

LEMLL	&\bf{0.969(1)}	&0.694(4)	&\bf{0.989(1)}	&\bf{0.980(1)}	&\bf{0.983(1)}	&\bf{0.983(1)}	&\bf{0.982(1)}	&\bf{0.975(1)}	&\bf{0.980(1)}	&\bf{0.986(1)}	&\bf{0.985(1)}	&\bf{0.987(1)}	&0.965(2)	&\bf{0.941(1)}	&0.857(2)	&1.333\cr
MLFE	&0.958(2)	&\bf{0.754(1)}	&0.957(4)	&0.937(2)	&0.907(2)	&0.917(2)	&0.915(2)	&0.923(2)	&0.909(2)	&0.910(2)	&0.910(2)	&0.914(2)	&\bf{0.967(1)}	&0.928(2)	&\bf{0.876(1)}	&1.933\cr
ML$^2$	&0.936(3)	&0.749(2)	&0.949(5)	&0.908(4)	&0.852(5)	&0.854(4)	&0.803(5)	&0.808(3)	&0.743(5)	&0.689(5)	&0.672(5)	&0.642(5)	&0.942(3)	&0.879(3)	&0.775(3)	&4.000\cr
RELIAB	&0.892(4)	&0.718(3)	&0.984(2)	&0.936(3)	&0.896(4)	&0.890(3)	&0.836(4)	&0.782(4)	&0.752(4)	&0.814(4)	&0.773(4)	&0.751(4)	&0.901(4)	&0.812(4)	&0.563(5)	&3.733\cr
FCM	&0.856(5)	&0.527(6)	&0.977(3)	&0.861(5)	&0.905(3)	&0.820(5)	&0.842(3)	&0.723(5)	&0.803(3)	&0.895(3)	&0.843(3)	&0.884(3)	&0.804(5)	&0.766(5)	&0.718(4)	&4.067\cr
KM	&0.637(6)	&0.621(5)	&0.811(6)	&0.695(6)	&0.559(6)	&0.586(6)	&0.492(6)	&0.525(6)	&0.458(6)	&0.429(6)	&0.420(6)	&0.384(6)	&0.675(6)	&0.716(6)	&0.500(6)	&5.933\cr

    \midrule
    \multirow{2}{0.7cm}{Algorithm}&
    \multicolumn{15}{c}{$\rho=0.2$}&\multirow{2}{0.7cm}{Avg. Rank}\cr
    \cmidrule(lr){2-16}
    &SJA&NS&spoem&spo5&dtt&cold&heat&spo&diau&elu&cdc&alpha&3DFE&Mov&HG \cr
    \midrule

LEMLL	&\bf{0.969(1)}	&0.695(4)	&\bf{0.989(1)}	&\bf{0.980(1)}	&\bf{0.983(1)}	&\bf{0.983(1)}	&\bf{0.984(1)}	&\bf{0.977(1)}	&\bf{0.977(1)}	&\bf{0.986(1)}	&\bf{0.983(1)}	&\bf{0.985(1)}	&0.964(2)	&\bf{0.941(1)}	&0.866(2)	&1.333\cr
MLFE	&0.959(2)	&0.755(2)	&0.957(4)	&0.937(2)	&0.907(2)	&0.917(2)	&0.922(2)	&0.928(2)	&0.918(2)	&0.907(3)	&0.908(2)	&0.906(3)	&\bf{0.968(1)}	&0.928(2)	&\bf{0.886(1)}	&2.133\cr
ML$^2$	&0.940(3)	&\bf{0.765(1)}	&0.949(5)	&0.908(4)	&0.852(5)	&0.854(4)	&0.844(5)	&0.842(4)	&0.829(5)	&0.752(5)	&0.731(5)	&0.732(5)	&0.944(3)	&0.879(3)	&0.824(3)	&4.000\cr
RELIAB	&0.897(4)	&0.722(3)	&0.984(2)	&0.936(3)	&0.896(4)	&0.890(3)	&0.909(4)	&0.852(3)	&0.850(3.5)	&0.882(4)	&0.847(4)	&0.865(4)	&0.906(4)	&0.812(4)	&0.684(5)	&3.633\cr
FCM	&0.860(5)	&0.526(6)	&0.977(3)	&0.861(5)	&0.905(3)	&0.820(5)	&0.912(3)	&0.790(5)	&0.850(3.5)	&0.927(2)	&0.881(3)	&0.940(2)	&0.808(5)	&0.766(5)	&0.766(4)	&3.967\cr
KM	&0.641(6)	&0.621(5)	&0.811(6)	&0.695(6)	&0.559(6)	&0.586(6)	&0.579(6)	&0.583(6)	&0.609(6)	&0.520(6)	&0.507(6)	&0.516(6)	&0.686(6)	&0.716(6)	&0.608(6)	&5.933\cr

    \midrule
    \multirow{2}{0.7cm}{Algorithm}&
    \multicolumn{15}{c}{$\rho=0.3$}&\multirow{2}{0.7cm}{Avg. Rank}\cr
    \cmidrule(lr){2-16}
    &SJA&NS&spoem&spo5&dtt&cold&heat&spo&diau&elu&cdc&alpha&3DFE&Mov&HG \cr
    \midrule

LEMLL	&\bf{0.973(1)}	&0.713(4)	&\bf{0.989(1)}	&\bf{0.980(1)}	&\bf{0.987(1)}	&\bf{0.985(1)}	&\bf{0.986(1)}	&\bf{0.982(1)}	&\bf{0.977(1)}	&\bf{0.985(1)}	&\bf{0.984(1)}	&\bf{0.984(1)}	&0.964(2)	&\bf{0.942(1)}	&0.871(2)	&1.333\cr
MLFE	&0.965(2)	&0.784(3)	&0.957(4)	&0.937(2)	&0.931(4)	&0.937(3)	&0.925(4)	&0.936(2)	&0.918(2)	&0.911(4)	&0.913(3)	&0.910(3)	&\bf{0.970(1)}	&0.931(2)	&\bf{0.892(1)}	&2.667\cr
ML$^2$	&0.957(3)	&\bf{0.810(1)}	&0.949(5)	&0.908(4)	&0.908(5)	&0.907(4)	&0.871(5)	&0.884(4)	&0.834(5)	&0.794(5)	&0.795(5)	&0.78(5)	&0.959(3)	&0.887(3)	&0.854(3)	&4.000\cr
RELIAB	&0.953(4)	&0.802(2)	&0.984(2)	&0.936(3)	&0.966(2)	&0.946(2)	&0.943(2)	&0.930(3)	&0.855(3)	&0.918(3)	&0.911(4)	&0.907(4)	&0.943(4)	&0.824(4)	&0.761(5)	&3.133\cr
FCM	&0.835(5)	&0.563(6)	&0.977(3)	&0.861(5)	&0.963(3)	&0.880(5)	&0.935(3)	&0.876(5)	&0.854(4)	&0.942(2)	&0.919(2)	&0.952(2)	&0.836(5)	&0.771(5)	&0.788(4)	&3.933\cr
KM	&0.749(6)	&0.630(5)	&0.811(6)	&0.695(6)	&0.734(6)	&0.713(6)	&0.659(6)	&0.686(6)	&0.623(6)	&0.594(6)	&0.609(6)	&0.593(6)	&0.759(6)	&0.733(6)	&0.682(6)	&5.933\cr

    \midrule
    \multirow{2}{0.7cm}{Algorithm}&
    \multicolumn{15}{c}{$\rho=0.4$}&\multirow{2}{0.7cm}{Avg. Rank}\cr
    \cmidrule(lr){2-16}
    &SJA&NS&spoem&spo5&dtt&cold&heat&spo&diau&elu&cdc&alpha&3DFE&Mov&HG \cr
    \midrule

LEMLL	&\bf{0.971(1)}	&0.725(4)	&\bf{0.989(1)}	&\bf{0.986(1)}	&\bf{0.988(1)}	&\bf{0.988(1)}	&\bf{0.988(1)}	&\bf{0.984(1)}	&\bf{0.978(1)}	&\bf{0.987(1)}	&\bf{0.986(1)}	&\bf{0.985(1)}	&\bf{0.960(1)}	&0.955(2)	&0.874(2)	&1.333\cr
MLFE	&0.959(2)	&0.801(3)	&0.957(4)	&0.959(3)	&0.934(4)	&0.944(3)	&0.940(4)	&0.946(3)	&0.935(2)	&0.923(4)	&0.922(4)	&0.919(4)	&0.959(2)	&\bf{0.959(1)}	&\bf{0.896(1)}	&2.933\cr
ML$^2$	&0.953(4)	&0.831(2)	&0.949(5)	&0.946(4)	&0.914(5)	&0.924(4)	&0.907(5)	&0.912(5)	&0.884(4)	&0.847(5)	&0.839(5)	&0.828(5)	&0.954(3)	&0.939(3)	&0.872(3)	&4.133\cr
RELIAB	&0.954(3)	&\bf{0.844(1)}	&0.984(2)	&0.979(2)	&0.970(2)	&0.964(2)	&0.971(2)	&0.959(2)	&0.904(3)	&0.950(3)	&0.944(2)	&0.938(3)	&0.952(4)	&0.898(4)	&0.812(4)	&2.600\cr
FCM	&0.857(5)	&0.600(6)	&0.977(3)	&0.943(5)	&0.964(3)	&0.908(5)	&0.967(3)	&0.913(4)	&0.867(5)	&0.951(2)	&0.942(3)	&0.960(2)	&0.877(5)	&0.832(6)	&0.801(5)	&4.133\cr
KM	&0.774(6)	&0.635(5)	&0.811(6)	&0.805(6)	&0.759(6)	&0.779(6)	&0.749(6)	&0.752(6)	&0.735(6)	&0.693(6)	&0.686(6)	&0.675(6)	&0.784(6)	&0.850(5)	&0.737(6)	&5.867\cr

    \midrule
    \multirow{2}{0.7cm}{Algorithm}&
    \multicolumn{15}{c}{$\rho=0.5$}&\multirow{2}{0.7cm}{Avg. Rank}\cr
    \cmidrule(lr){2-16}
    &SJA&NS&spoem&spo5&dtt&cold&heat&spo&diau&elu&cdc&alpha&3DFE&Mov&HG \cr
    \midrule

LEMLL	&\bf{0.967(1)}	&0.731(4)	&\bf{0.989(1)}	&\bf{0.985(1)}	&\bf{0.988(1)}	&\bf{0.988(1)}	&\bf{0.989(1)}	&\bf{0.985(1)}	&\bf{0.983(1)}	&\bf{0.988(1)}	&\bf{0.988(1)}	&\bf{0.987(1)}	&\bf{0.949(1)}	&0.955(2)	&0.875(2)	&1.333\cr
MLFE	&0.957(3)	&0.809(3)	&0.957(4)	&0.969(3)	&0.934(4)	&0.944(3)	&0.943(4)	&0.952(3)	&0.948(2)	&0.935(4)	&0.934(4)	&0.932(4)	&0.946(2)	&\bf{0.962(1)}	&\bf{0.898(1)}	&3.000\cr
ML$^2$	&0.954(4)	&0.840(2)	&0.950(5)	&0.961(4)	&0.914(5)	&0.924(4)	&0.917(5)	&0.927(5)	&0.915(4)	&0.880(5)	&0.874(5)	&0.869(5)	&0.942(3)	&0.947(3)	&0.873(3)	&4.133\cr
RELIAB	&0.960(2)	&\bf{0.866(1)}	&0.985(2)	&0.982(2)	&0.970(2)	&0.964(2)	&0.975(2)	&0.974(2)	&0.941(3)	&0.965(2)	&0.964(2)	&0.958(3)	&0.937(4)	&0.914(4)	&0.846(4)	&2.467\cr
FCM	&0.863(5)	&0.621(6)	&0.978(3)	&0.941(5)	&0.964(3)	&0.909(5)	&0.972(3)	&0.941(4)	&0.882(5)	&0.954(3)	&0.953(3)	&0.965(2)	&0.868(5)	&0.847(6)	&0.806(5)	&4.200\cr
KM	&0.827(6)	&0.638(5)	&0.812(6)	&0.882(6)	&0.759(6)	&0.779(6)	&0.779(6)	&0.800(6)	&0.798(6)	&0.758(6)	&0.754(6)	&0.751(6)	&0.811(6)	&0.880(5)	&0.780(6)	&5.867\cr

    \midrule
    \midrule
    \end{tabular}
\end{table*}

Based on the experimental results, the following observations can be apparently made:

\begin{itemize}
\item LEMLL achieves optimal (lowest) average rank in terms of each evaluation metric (Table IV). On the 15 benchmark data sets, across all the evaluation metrics, LEMLL ranks 1st in 69.3\% cases and ranks 2nd in 21.3\% cases.
\item When compared with the three well-established two-step approaches, on the 15 data sets (Table II) (Table III), across all the evaluation metrics, LEMLL is significantly superior to MLFE in 89.3\% cases, LEMLL is significantly superior to ML$^2$ in 90.7\% cases and LEMLL is significantly superior to RELIAB in 80\% cases. Thus LEMLL achieves superior performance over those two-stage approaches.
\item When compared with the three state-of-the-art algorithms, on the 15 data sets (Table II) (Table III), across all the evaluation metrics, LEMLL is significantly superior to ML-kNN in 82.7\% cases, LEMLL is significantly superior to CLR in 93.3\% cases and LEMLL is significantly superior to ECC in 88\% cases.
\item Another interesting observation is that on all the data sets (Table II) (Table III), across all the evaluation metrics, the performance of LEMLL is superior or equal to MSVR, and LEMLL is significantly superior to MSVR in 76\% cases, which verify the superiority of the reconstructed numerical labels to the logical labels.
\end{itemize}

To summarize, LEMLL achieves superior performance over the well-established two-stage algorithms and the three state-of-the-art algorithms across extensive benchmark data sets. LEMLL significantly outperforms MSVR in most cases, which validates the effectiveness of label enhancement for boosting the multi-label learning performance.

\subsection{Reconstruction Performance Evaluation}
\subsubsection{Experimental Settings}
To further evaluate the numerical labels $\bm{\mu}$ reconstructed by LEMLL, experimental studies on 15 real-world label distribution data sets \cite{geng2016label} with ground-truth label importance are conducted. Table V summarizes the detailed characteristics of the 15 real-world data sets.

Note that the problem of reconstructing label importance from logical labels is relatively new, and the logical multi-label data with ground-truth label importance is not available yet. Thus we consider the following settings of the reconstruction tasks. In a label distribution data set, each instance is associated with a label distribution. The data set used in our experiments, however, contains for each instance not the real distribution, but a set of labels. The set includes the labels with the highest weights in the distribution, and is the smallest set such that the sum of these weights exceeds a given threshold. The settings can model, for instance, the way in which annotators label images or add keywords to texts: it assumes that annotators add labels starting with the most relevant ones, until they feel the labeling is sufficiently complete. Therefore, the logical labels in the data sets can be binarized from the real label distributions as follows. For each instance $\bm{x}$, of which the label distribution is $\bm{d}=[d_{\bm{x}}^{y_{1}},d_{\bm{x}}^{y_{2}},...,d_{\bm{x}}^{y_{l}}]^T$, the greatest description degree $d_{\bm{x}}^{y_{j}}$ is found, and the label $y_{j}$ is set to relevant label. Then, we calculate the sum of the description degrees of all the current relevant labels $H = \sum_{y_{j} \in S^{re}} d_{\bm{x}}^{y_{j}}$, where $S^{re}$ is the set of the current relevant labels. If $H$ is less than a predefined threshold $\rho$, we continue finding the greatest description degree among other labels excluded from $S^{re}$ and select the label corresponding to the greatest description degree into $S^{re}$. This process continues until $H > \rho$. Finally, the logical labels to the labels in $S^{re}$ are set to 1, and other logical labels are set to $-1$. In the experiments, $\rho$ varies from 0.1 to 0.5 with step size of 0.1. Thus we use each label distribution data set to form five logical multi-label data sets.

After binarizing the logical labels from the ground-truth label distributions, we recover the numerical labels from the logical labels via the LE algorithms and then the numerical labels are transferred
to the label distribution via normalization $g(\bm{\mu}) = sigmoid(\bm{\mu})/Z$, where $sigmoid(\bm{\cdot})$ is the sigmoid function mapping numerical value into $(0,1)$ and $Z$ is the normalization factor, i.e., $Z = \sum_{j=1}^lsigmoid(\mu_j)$. Finally, we compare the reconstructed label distributions with the ground-truth label distributions.

In order to evaluate the similarity between the reconstructed label distributions and the ground-truth label distributions, as suggested in \cite{geng2016label}, three measures are chosen for our experiments, which include Chebyshev distance (Chebyshev), Kullback-Leibler divergence (K-L) and cosine coefficient (Cosine). The first two are distance measures and the smaller the values the better the performance. The last one is similarity measures and the larger the values the better the performance.

We choose to compare the performance of LEMLL against the five LE algorithms mentioned in Section II, i.e., FCM \cite{el2006study}, KM \cite{jiang2006fuzzy} and the first stage of MLFE \cite{zhang2018feature}, ML$^2$ \cite{hou2016multi} and RELIAB \cite{li2015leveraging}. For each comparing approaches, the parameters recommended in the corresponding literatures are used. For MLFE,  the penalty parameter $\rho$ is set to 1, $c_1$ is set to 1 and $c_2$ is set to 2. For ML$^2$, $K$ is set to $l$ +1 and $\lambda$ is set to 1. For RELIAB, $\alpha$ is set to 0.5. For FCM, $\beta$ is set to 2. For LEMLL, $K$ is set to 10. $\varepsilon$ is set to 0.1. $\alpha$, $\beta$ and $\gamma$ are all set to 1. Linear kernel is used in KM and LEMLL.

\subsubsection{Experimental Results}
We run the LEMLL, FCM, KM and the first stage of MLFE, ML$^2$ and RELIAB with threshold $\rho$ varying from 0.1 to 0.5 with step size of 0.1 on the 15 data sets. Table VI, Table VII and Table VIII report the results of the six LE algorithms across all the threshold $\rho$ on all the data sets evaluated by  Chebyshev, K-L and Cosine respectively. The best reconstruction performance on each measure is highlighted by boldface and average ranks are given in the last column. Note that this experiment is a \emph{reconstruction task}, not a predictive task. Thus each LE algorithm only runs once. As shown in Table VI, Table VII and Table VIII, the following observation can be made: Across all the threshold $\rho$, LEMLL achieves optimal (lowest) average rank in terms of each evaluation metric. On the 15 data sets, across all the threshold $\rho$, across all the evaluation metrics, LEMLL ranks 1st in 80.4\% cases and ranks 2nd in 11.1\% cases.

To summarize, LEMLL achieves superior reconstruction performance over other algorithms, which demonstrates that LEMLL has good capability in reconstructing latent label importance information from logical multi-label data.

\section{Conclusion}
This paper proposes a framework of multi-label learning with label enhancement. Extensive comparative studies clearly validate the performance of multi-label learning can be improved significantly with label enhancement and LEMLL can effectively reconstruct latent label importance information from logical multi-label data. In the future, we will explore if there are other assumptions about label enhancement.

\section*{Acknowledgments}
This research was supported by the National Key Research $\&$ Development Plan of China (No. 2017YFB1002801), the National Science Foundation of China (61622203), the Jiangsu Natural Science Funds for Distinguished Young Scholar (BK20140022), the Collaborative Innovation Center of Novel Software Technology and Industrialization, and the Collaborative Innovation Center of Wireless Communications Technology.

\bibliographystyle{IEEEtran}
\bibliography{main}

\end{document}